\documentclass[runningheads]{llncs}
\usepackage{graphicx}

\usepackage{tikz}
\usepackage{comment} 
\usepackage{amsmath,amssymb} 
\usepackage{color}

\usepackage{url}
\usepackage{booktabs}
\usepackage{multirow}
\usepackage{pifont}
\usepackage{marvosym}

\begin{document}
\pagestyle{headings}
\mainmatter
\def\ECCVSubNumber{3807}  

\title{PSConv: Squeezing Feature Pyramid into One Compact Poly-Scale Convolutional Layer} 

\titlerunning{PSConv: A Compact Poly-Scale Convolutional Layer}
%
\author{Duo Li\inst{1,2}\thanks{indicates intern at Intel Labs China. \textsuperscript{\Letter} indicates corresponding authors.} \and
	Anbang Yao\inst{2}\textsuperscript{\Letter} \and
	Qifeng Chen\inst{1}\textsuperscript{\Letter}}
\authorrunning{D. Li, A. Yao and Q. Chen}
%
\institute{The Hong Kong University of Science and Technology \and
	Intel Labs China \\
	\email{duo.li@connect.ust.hk}\qquad
	\email{anbang.yao@intel.com}\qquad
	\email{cqf@ust.hk}}
\maketitle

\begin{abstract}
	Despite their strong modeling capacities, Convolutional Neural Networks (CNNs) are often scale-sensitive. For enhancing the robustness of CNNs to scale variance, multi-scale feature fusion from different layers or filters attracts great attention among existing solutions, while the more granular kernel space is overlooked. We bridge this regret by exploiting multi-scale features in a finer granularity. The proposed convolution operation, named Poly-Scale Convolution (PSConv), mixes up a spectrum of dilation rates and tactfully allocate them in the individual convolutional kernels of each filter regarding a single convolutional layer. Specifically, dilation rates vary cyclically along the axes of input and output channels of the filters, aggregating features over a wide range of scales in a neat style. PSConv could be a drop-in replacement of the vanilla convolution in many prevailing CNN backbones, allowing better representation learning without introducing additional parameters and computational complexities. Comprehensive experiments on the ImageNet and MS COCO benchmarks validate the superior performance of PSConv. Code and models are available at \url{https://github.com/d-li14/PSConv}.
\end{abstract}

\section{Introduction}

With the booming development of CNNs, dramatic progress has been made in the field of computer vision. As an inherent feature extraction mechanism, CNNs naturally learn coarse-to-fine hierarchical image representations. To mimic human visual systems that could process instances and stuff concurrently, it is of vital importance for CNNs to gather diverse information from objects of various sizes and understand meaningful contextual backgrounds. However, streamlined CNNs usually have fixed-sized receptive fields, lacking the ability to tackle this kind of issue. Such a deficiency restricts their performance on visual recognition tasks, especially scale-sensitive dense prediction problems. The advent of FCN~\cite{Long_2015_CVPR} and Inception~\cite{Szegedy_2015_CVPR} demonstrates the privilege of multi-scale representation to perceive heterogeneous receptive fields with impressive performance improvement. Motivated by these pioneering works, follow-up approaches explore and upgrade multi-scale feature fusion with more intricate skip connections or parallel streams. However, we notice that most of the existing works capture these informative multi-scale features in a layer-wise or filter-wise style, laying emphasis on the architecture engineering of the entire network or their composed building blocks.

From a brand new perspective, we shift the focus of design from macro- to micro-architecture towards the target of easily exploiting multi-scale features without touching the overall network architecture. Expanding kernel sizes and extending the sampling window sizes via increasing dilation rates are two popular techniques to enlarge the receptive fields inside one convolution operation. Compared to large kernels that bring about more parameter storage and computational consumption, dilated convolution is an alternative to cope with objects in an array of scales without introducing extra computational complexities. In this paper, we present Poly-Scale Convolution (PSConv), a novel convolution operation, extracting multi-scale features from the more granular convolutional kernel space. PSConv respects two design principles: firstly, regarding one single convolutional filter, its constituent kernels use a group of dilation rates to extract features corresponding to different receptive fields; secondly, regarding all convolutional filters in one single layer, the group of dilation rates corresponding to each convolutional filter alternates along the axes of input and output channels in a cyclic fashion, extracting diverse scale information from the incoming features and mapping them into outgoing features in a wide range of scales. Through these atomic operations on individual convolutional kernels, we effectively dissolve the aforementioned deficiency of standard convolution and push the multi-scale feature fusion process to a much more granular level. This proposed approach tiles the \textit{kernel lattice}\footnote{\textit{kernel lattice} refers to the two-dimensional flattened view of convolutional filters where the kernel space is reduced while the channel space is retained, thus each cell in the lattice represents an individual kernel (see Fig.~\ref{fig:conv} for intuitive illustration).} with hierarchically stacked pyramidal features defined in the previous methodologies~\cite{Lin_2017_CVPR}. Each specific feature scale in one pyramid layer can be grasped with a collection of convolutional kernels in a PSConv operation with the same corresponding dilation rate, thus the whole feature pyramid can be represented in a condensed fashion using one compact PSConv layer with a spectrum of dilation rates. \textit{Poly-Scale} Convolution extends the conventional \textit{mono-scale} convolution living on a homogeneous dilation space of kernel lattice, hence the name of this convolution form. In our PSConv, scale-aware features located in different channels collaborate as a unity to deal with scale variance problems, which is critical for handling a single instance with a non-rigid shape or multiple instances with complex scale variations. For scale-variant stimuli, PSConv is capable of learning self-adaptive attention for different receptive fields following a dynamic routing mechanism, improving the representation ability without any additional parameters or memory cost.

Thanks to its plug-and-play characteristic, our PSConv can be readily used to replace the vanilla convolution of arbitrary state-of-the-art CNN architectures, \textit{e}.\textit{g}., ResNet~\cite{He_2016_CVPR}, giving rise to PS-ResNet. We also build PS-ResNeXt featuring group convolutions to prove the universality of PSConv. These models are comprehensively evaluated on the ImageNet~\cite{imagenet_cvpr09} dataset and show consistent gains over the baseline of plain CNN counterparts. More experiments on (semi-)dense prediction tasks, \textit{e}.\textit{g}., object detection and instance segmentation on the MS COCO dataset, further demonstrate the superiority of our proposed PSConv over the standard ones under the circumstances with severe scale variations. It should be noted that PSConv is also independent of other macro-architectural choices and thus orthogonal and complementary to existing multi-scale network designs at a coarser granularity, leaving extra room to combine them together for further performance enhancement.

Our core contributions are summarized as follows:
\begin{description}
	\item[\ding{111}] We extend the scope of the conventional mono-scale convolution operation by developing our Poly-Scale Convolution, which effectively and efficiently aggregates multi-scale features via arranging a spectrum of dilation rates in a cyclic manner inside the kernel lattice.
	\item[\ding{111}] We investigate the multi-scale network design through the lens of kernel engineering instead of network engineering, which avoids the necessity of tuning network structure or layer configurations while achieves competitive performance, when adapted to existing CNN architectures.
\end{description}

\section{Related Work}

We briefly review previous relevant network and modular designs and clarify their similarities and differences compared to our proposed approach.

\textbf{Multi-Scale Network Design.} Early works like AlexNet~\cite{NIPS2012_4824} and VGGNet~\cite{vgg} learn multi-scale features in a data-driven manner, which are naturally equipped with a hierarchical representation by the inherent design of CNNs. The shallow layers seek finer structures in the images like edges, corners, and texture, while deep layers abstract semantic information, such as outlines and categories. In order to break the limitation of fixed-sized receptive fields and enhance feature representation, many subsequent works based on explicit multi-scale feature fusion are presented. Within this scope, there exists a rich literature making innovations on \textbf{skip connection} and \textbf{parallel stream}.

The \textbf{skip connection} structure exploits features with multi-size receptive fields from network layers at different depths. The representative FCN~\cite{Long_2015_CVPR} adds up feature maps from multiple intermediate layers with the skip connection. Analogous techniques have also been applied to the field of edge detection, presented by HED~\cite{Xie_2015_ICCV}. In the prevalent encoder-decoder architecture, the decoder network could be a symmetric version of the encoder network, with skip connections over some mirrored layers~\cite{Noh_2015_ICCV} or concatenation of feature maps~\cite{10.1007/978-3-319-24574-4_28}. DLA~\cite{Yu_2018_CVPR} extends the peer-to-peer skip connections into a tree structure, aggregating features from different layers in an iterative and hierarchical style. FishNet~\cite{NIPS2018_7356} stacks an upsampling body and a downsampling head upon the backbone tail, refining features that compound multiple resolutions.

The \textbf{parallel stream} structure generates multi-branch features conditioned on a spectrum of receptive fields. Though too numerous to list in full, recent research efforts often attack conventional designs via either maintaining a feature pyramid virtually from bottom to top or repeatedly stacking split-transform-merge building blocks. The former pathway of design includes several exemplars like Multigrid~\cite{Ke_2017_CVPR} and HRNet~\cite{2019arXiv190404514S}, which operate on a stack of features with different resolutions in each layer. Similarly, Octave Convolution~\cite{chen2019drop} decomposes the standard convolution into two resolutions to process features at different frequencies, removing spatial redundancy by separating scales. The latter pathway of design is more crowded with the following works. The Inception~\cite{Szegedy_2015_CVPR,pmlr-v37-ioffe15,Szegedy_2016_CVPR} family utilizes parallel pathways with various kernel sizes in one Inception block. BL-Net~\cite{chen2018biglittle} is composed of branches with different computational complexities, where the features at the larger scale pass through fewer blocks to spare computational resources and the features from different branches at distinct scales are merged with a linear combination. Res2Net~\cite{gao2019res2net} and OSNet~\cite{Zhou_2019_ICCV} construct a group of hierarchical residual-like connections or stacked Lite $3 \times 3$ layers along the channel axis in one single residual block. ELASTIC~\cite{Wang_2019_CVPR} and ScaleNet~\cite{Li_2019_CVPR} learn a soft scaling policy to allocate weights for different resolutions in the paratactic branches. Despite distinct with respect to detailed designs, these works all extensively use down-sampling or up-sampling to resize the features to $2^n$ times and inevitably adjust the original architecture via the selection of new hyperparameters and layer configurations when plugged in. On the contrary, our proposed PSConv can be a straightforwardly drop-in replacement of the vanilla convolution, leading a trend towards more effective and efficient multi-scale feature representation. Conventionally, features with multi-size receptive fields are integrated via channel concatenation, weighted summation or attention models. In stark contrast, we suggest to explore multi-scale features in a finer granularity, encompassed in merely one single convolutional layer.

In addition to the aforementioned networks designed to enhance image classification, scale variance poses more challenges in (semi-)dense prediction tasks, \textit{e}.\textit{g}., object detection and semantic segmentation. Faster R-CNN~\cite{Girshick_2015_ICCV} uses pre-defined anchor boxes of different sizes to address this issue. DetNet~\cite{Li_2018_ECCV}, RFBNet~\cite{Liu_2018_ECCV} and TridentNet~\cite{Li_2019_ICCV} apply dilated convolutions to enlarge the receptive fields. DeepLab~\cite{Chen_2018_ECCV} and PSPNet~\cite{Zhao_2017_CVPR} construct feature pyramid in a parallel fashion. FPN~\cite{Lin_2017_CVPR} is designed to fuse features at multiple resolutions through top-down and lateral connections and provides anchors specific to different scales.

\textbf{Dynamic Convolution.} All approaches above process multi-scale information without drilling down into the pure single convolutional layer. Complementarily, another line of research concentrates on injecting scale modules into the original network directly and handling various receptive fields in an automated fashion. STN~\cite{NIPS2015_5854} explicitly learns a parametric manipulation of the feature map conditioned on itself to improve the tolerance to spatial geometric transformations. ACU~\cite{Jeon_2017_CVPR} and DCN~\cite{Dai_2017_ICCV,Zhu_2019_CVPR} learn offsets at each sampling position of the convolutional kernel or the feature map to permit a flexible shape deformation during the convolution process. SAC~\cite{Zhang_2017_ICCV} inserts an extra regression layer to densely infer the scale coefficient map and applies an adaptive dilation rate to the convolutional kernel at each spatial location of the feature map. POD~\cite{Peng_2019_ICCV} predicts a globally continuous scale and then converts the learned fractional scale to a channel-wise combination of integer scales for fast deployment. We respect the succinctness of these plugged-in modules and follow these approaches in their form. In this spirit, we formulate a novel convolution representation through cyclically alternating dilation rates along both input and output channel dimensions to address the scale variations. We also note that some of the aforementioned modules are designed specifically for (semi-)dense prediction problems, \textit{e}.\textit{g}., SAC, DCN, and POD and others do not scale to large-scale classification benchmarks like ImageNet, \textit{e}.\textit{g}., STN for MNIST and SVHN, ACU for CIFAR. In contrast, our proposed PSConv focuses on backbone engineering, empirically shows its effectiveness on ImageNet and generalizes well to other complicated tasks on MS COCO. Furthermore, while the offsets are learned efficiently in some methods (ACU, SAC, and DCN), the inference is time-consuming due to the dynamic grid sampling and the bilinear interpolation at each position. Aligning to the tenet of POD~\cite{Peng_2019_ICCV}, it is unnecessary to permit too much freedom with floating-point offsets at each spatial location as DCN~\cite{Dai_2017_ICCV} and learning in such an aggressive manner places an extra burden on the inference procedure. We opt for a better accuracy-efficiency trade-off by constraining dilation rates in the integer domain and organizing them into repeated partitions. Last but not least, the recently proposed MixConv~\cite{Tan_2019_BMVC} may be the most related scale module compared to PSConv, which will be discussed at the end of the next section.

\section{Method}
\label{sec:method}

Compared to previous multi-scale feature fusion solutions in a coarse granularity, we seek an alternative design with the finer granularity and stronger feature extraction ability, while maintaining a similar computational load.

\subsection{Sketch of Convolution Operations}

We initiate from elaborating the vanilla (dilated) convolution process to make the definition of our proposed PSConv self-contained. For a single convolutional layer, let the tensor $\mathcal{F} \in \mathbb{R}^{C_{in} \times H \times W}$ denotes its input feature map with the shape of $C_{in} \times H \times W$, where $C_{in}$ is the number of channels, $H$ and $W$ are the height and width respectively. A set of $C_{out}$ filters with the kernel size $K \times K$ are convolved with the input tensor individually to obtain the desired output feature map with $C_{out}$ channels, where each filter has $C_{in}$ kernels to match those channels in the input feature map. Denote the above filters as $\mathcal{G} \in \mathbb{R}^{C_{out} \times C_{in} \times K \times K}$, then the vanilla convolution operation can be represented as
\begin{equation}
\mathcal{H}_{c,x,y} = \sum_{k=1}^{C_{in}} \sum_{i=-\frac{K-1}{2}}^{\frac{K-1}{2}} \sum_{j=-\frac{K-1}{2}}^{\frac{K-1}{2}} \mathcal{G}_{c,k,i,j} \mathcal{F}_{k,x+{\color{red}i},y+{\color{red}j}},
\end{equation}
\begin{figure}[htbp]
	\begin{center}
		\includegraphics[width=.9\linewidth,trim=5 5 5 5,clip]{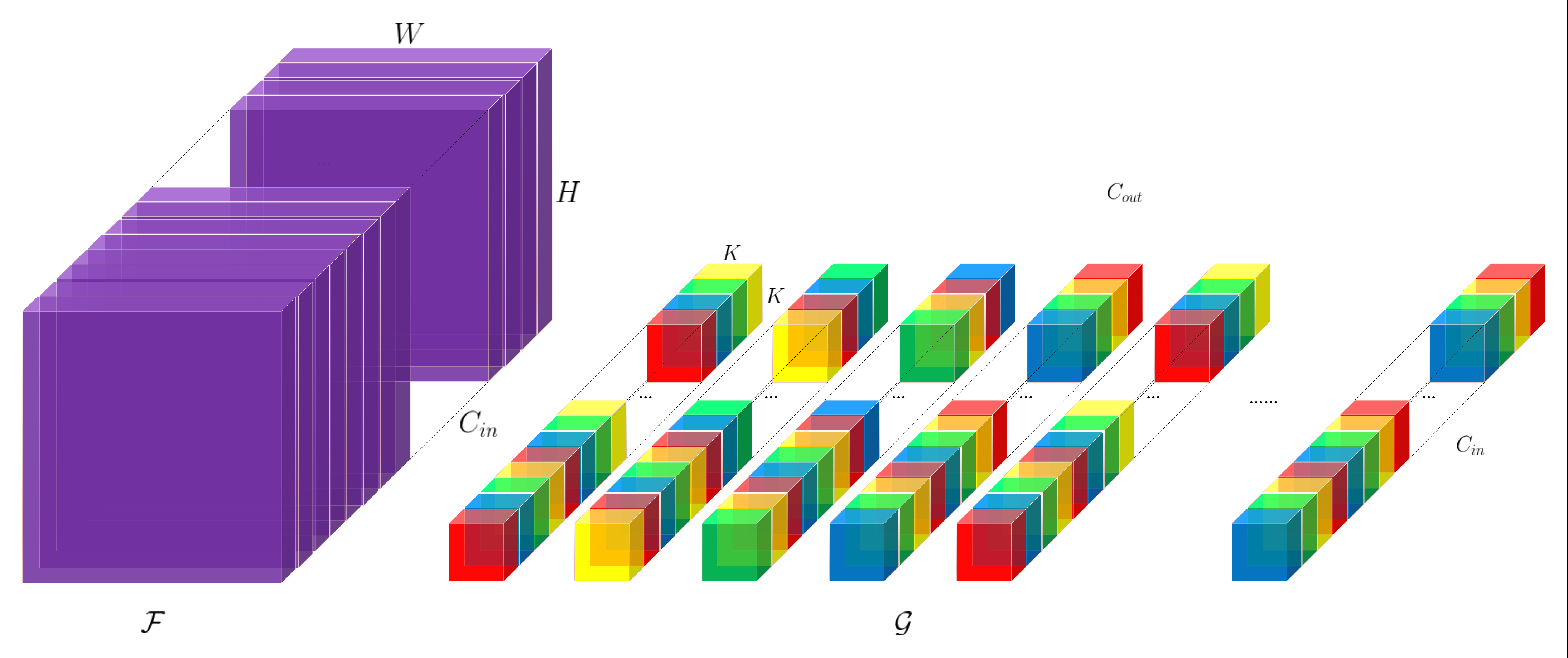}
	\end{center}
	\vskip -0.3in
	\caption{Schematic illustration of our proposed PSConv operation. $\mathcal{F}$ represents the input feature map and $\mathcal{G}$ represents $C_{out}$ convolutional filters in a set. Convolutional kernels with the same dilation rates in the set of filters $\mathcal{G}$ are rendered with the same color. Best viewed in color.}
	\label{fig:teaser}
	\vskip -0.2in
\end{figure}where $\mathcal{H}_{c,x,y}$ is one element in the output feature map  $\mathcal{H} \in \mathbb{R}^{C_{out} \times H \times W}$, $c = 1, 2, \cdots, C_{out}$ is the index of an output channel, $x = 1, 2, \cdots, H$ and $y = 1, 2, \cdots, W$ are indices of spatial positions in the feature map.

Dilated Convolution~\cite{Yu_2017_CVPR} enlarges sampling intervals in the spatial domain to cover objects of larger sizes. A dilated convolution with the dilation rate $d$ can be represented as
\begin{equation}
\mathcal{H}_{c,x,y} = \sum_{k=1}^{C_{in}} \sum_{i=-\frac{K-1}{2}}^{\frac{K-1}{2}} \sum_{j=-\frac{K-1}{2}}^{\frac{K-1}{2}} \mathcal{G}_{c,k,i,j} \mathcal{F}_{k,x+{\color{red}id},y+{\color{red}jd}}.
\end{equation}

Noticing that a combination of dilation rates is conducive to extract both global and local information, we propose a new convolution form named Poly-Scale Convolution (PSConv) which scatters organized dilation rates over different kernels inside one convolutional filter. Furthermore, our PSConv integrates multi-scale features in a one-shot manner and brings characteristics of the dilated convolution into full play, thus without introducing additional computational cost.
To gather multi-scale information from different input channels via a linear summation, dilation rates are varied at different kernels in one convolutional filter. To process an input channel with various receptive fields, dilation rates are also varied in different filters for a certain channel. It is written as
\begin{equation}
\mathcal{H}_{c,x,y} = \sum_{k=1}^{C_{in}} \sum_{i=-\frac{K-1}{2}}^{\frac{K-1}{2}} \sum_{j=-\frac{K-1}{2}}^{\frac{K-1}{2}} \mathcal{G}_{c,k,i,j} \mathcal{F}_{k,x+{\color{red}iD_{(c,k)}},y+{\color{red}jD_{(c,k)}}},
\end{equation}
where $D \in \mathbb{R}^{C_{out} \times C_{in}}$ is a matrix composed of channel-wise and filter-wise dilation rates in two orthogonal dimensions. An element $D_{(c,k)}$ is associated with a specific channel in one filter to support $\mathcal{G}_{c,k,\cdot,\cdot}$ as a unique convolutional kernel, thus the whole matrix $D$ can be interpreted as a mathematical representation of the kernel lattice in its subspace of dilation rate.

\subsection{Design Details}

As stated above, our major work is to reformulate the dilation rate patterns in the subspace of kernel lattice. We ensure that each row and column of the matrix $D$ have non-identical elements to achieve the desired properties of multi-scale feature fusion. On the contrary, if we avoid and retain identical elements in one row, then we would not collect multi-scale information to produce a new output channel in this operation, and it can be boiled down to multi-stream transformation before concatenation; if the similar event occurs in one column, the corresponding input channel would not have necessarily diverse receptive fields covered, and it reduces to the split-transform-summation design of multi-scale networks. These are both suboptimal according to our ablative experiments in Table~\ref{tab:variation}. The illustration diagrams of these two simplified cases are provided in the supplementary materials.

\begin{figure}[htbp]
	\vskip -0.2in
	\begin{center}
		\includegraphics[width=\linewidth,trim=8 5 5 5,clip]{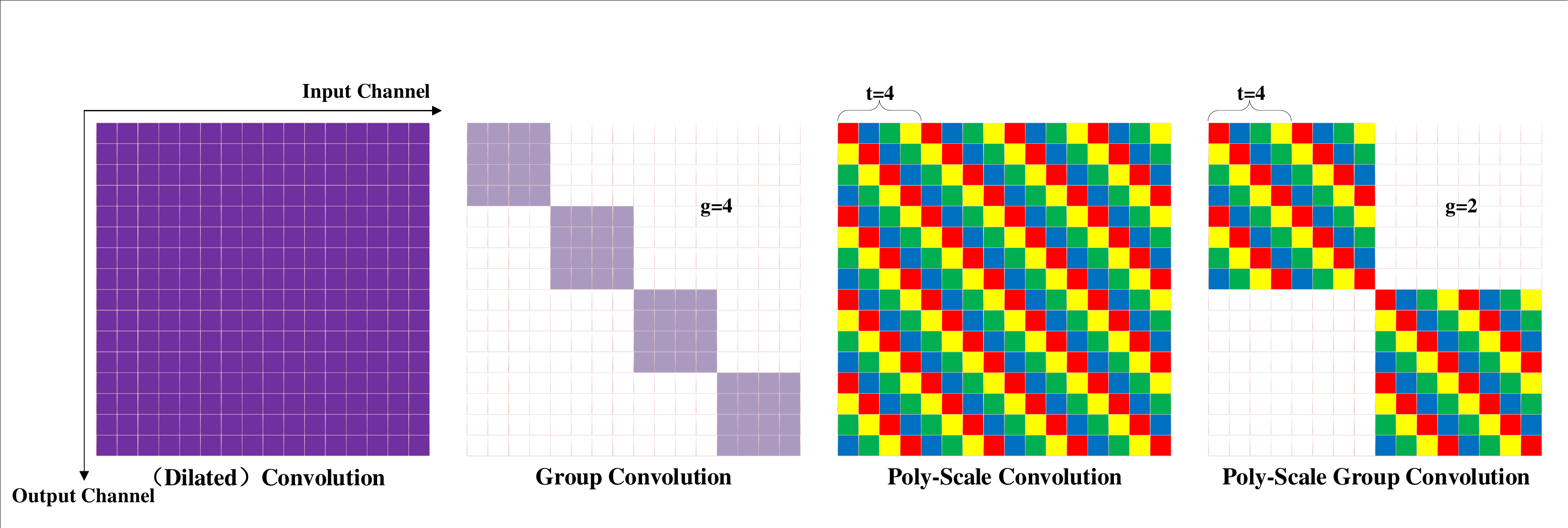}
	\end{center}
	\vskip -0.3in
	\caption{Comparison between dilation space of kernel lattice in different convolution operations. Kernels of standard convolution (with or without dilation) are showcased in the leftmost, where each kernel is located at one cell in the lattice. Group convolution (group number $g=4$) extensively utilized in the efficient network design is also included for reference. Poly-Scale convolution (cyclic interval $t=4$) and Poly-Scale group convolution (group number $g=2$ and cyclic interval $t=4$) in the right shows significant differences from the former two. Best viewed in color.}
	\label{fig:conv}
	\vskip -0.2in
\end{figure}

Following the above analysis, the design philosophy of PSConv could be decomposed into two coupled ingredients. Firstly, we concentrate on a single filter. In order to constrain the number of different dilation rates in a reasonable range, we heuristically arrange them inside one filter with a cyclic layout, \textit{i}.\textit{e}., dilation rates vary in a periodical manner along the axis of input channels. Specifically speaking, a total of $C_{in}$ input channels are divided into $P$ partitions. For each partition, $t=\lceil\frac{C_{in}}{P}\rceil$ channels are accommodated and a fixed pattern of dilation rates $\{d_1, d_2, \cdots, d_t\}$ is filled in to construct a row of the matrix $D$. Secondly, we broaden our horizons to all filters. In order to endow different filters with capacities to gather different kinds of scale combinations of input features, we adopt a shift-based strategy for dilation rates to flip the former filter to the latter one, \textit{i}.\textit{e}., the pattern of dilation rates regarding a convolutional filter is shifted by one channel to build its adjacent filter. In the illustrative example of Fig.~\ref{fig:conv}, $C_{in}=C_{out}=16$ and the partition number $P$ is set to 4, hence there leaves a blank of 4 dilation rates to be determined in the pattern $\{d_1, d_2, d_3, d_4\}$, where a specific colorization distinguishes one type of dilation rate from others. It is noted that viewed from the axis of output channels, dilation rates also present periodical variation. In other words, all types of dilation rates occur alternately along the vertical and horizontal axes in the trellis.

Furthermore, a comparison diagram is shown in Fig.~\ref{fig:conv}, to achieve better intuitive comprehension about different convolution operations. The filters of PSConv are exhibited from the vertical view of $\mathcal{G}$ (with appropriate rotate transformation) in Fig.~\ref{fig:teaser}, where each tile in the grid represents a kernel of $K \times K$ shape and the grid corresponds to the dilation rate matrix $D$. The filters of (dilated) convolution and group convolution are likewise displayed. In the conventional filters, if a dilation rate is applied, it will dominate the whole kernel lattice, while our PSConv has clear distinctions compared with them. We claim that merely varying dilation rates in the axis of output channels equals to using split-transform-merge units spanning a spectrum of dilation rates in the different streams. Our method takes one step further to spread the scale information along both input and output channel axes, pushing the selection of scale-variant features into the entire kernel space.
To the best of our knowledge, \textbf{it is the first attempt to mix up multi-scale information simultaneously in two orthogonal dimensions and leverage the complementary multi-scale benefits from such a fine granularity in the kernel space}.

It is noteworthy that PSConv is a generalized form of dilated convolution: since the cyclic interval $t$ decides how many types of dilation rates are contained in one partition, all kernels may share the same dilation rate once the partition number equals to that of input channels and then it degenerates into vanilla dilated convolution. The PSConv can also be applied to the group-wise convolution form by injecting the shared cyclic pattern into each group, as illustrated in the rightmost of Fig.~\ref{fig:conv}. Owing to the interchangeability of channel indices, grouping channels with the same dilation rate together leads to an equivalent but efficient implementation, which is depicted in the supplementary materials.

The recently proposed MixConv~\cite{Tan_2019_BMVC} might be similar to PSConv at the first glimpse. However, they are distinct regarding both the design principle and the application focus. On the one hand, MixConv integrates multiple kernel sizes for different patterns of resolutions which inevitably increases the parameters and computational budget, while PSConv mixes up a spectrum of dilation rates with a unified kernel size to economically condense multi-scale features within one convolution operation. Thus, for these two convolution forms, the manipulations on the kernel lattice are shaped from orthogonal perspectives. On the other hand, MixConv is dedicatedly developed for depthwise convolution (DWConv), while PSConv is versatile to both standard and group convolution. Due to the inherent constraint of DWConv, each individual channel in a MixConv operation exploits feature representation of a certain scale. However, in our PSConv, multi-scale representations are scattered along both input and output channels in a periodical manner. Hence, an individual channel could gather multifarious feature resolutions from the view of either input or output channels. We attach a more in-depth discussion around their differences and an illustration of the DWConv-based variant of PSConv in the supplementary materials.

\section{Experiments}

We conduct extensive experiments from conceptual to dense prediction tasks on several large-scale visual recognition benchmarks. Experimental results empirically validate the effectiveness and efficiency of our proposed convolution form. All experiments are performed with the PyTorch~\cite{NIPS2019_9015} library.

\subsection{ILSVRC 2012}
\label{sec:ilsvrc12}

ImageNet~\cite{imagenet_cvpr09} is one of the most challenging datasets for image classification, which is served as the benchmark of the ILSVRC2012 competition. It includes 1,281,167 training images and 50,000 validation images, and each image is manually annotated as one of the 1,000 object categories.

We incorporate our PSConv layer into various state-of-the-art convolutional neural networks, including ResNet~\cite{He_2016_CVPR}, ResNeXt~\cite{Xie_2017_CVPR} and SE-ResNet~\cite{Hu_2018_CVPR}. The training procedure is performed on the ImageNet training set by the SGD optimizer with the momentum of 0.9 and the weight decay of 1e-4. The mini-batch size is set to 256 and the optimization process lasts for a period of 120 epochs to achieve full convergence. The learning rate initiates from 0.1 and decays to zero following a half cosine function shaped schedule, the same as~\cite{chen2018biglittle} and~\cite{chen2019drop}. We adopt random scale and aspect ratio augmentation together with random horizontal flipping to process each training sample prior to feeding it into neural networks. We select the best-performing model along the training trajectory and report its performance on the ImageNet validation set. As is the common practice, we first resize the shorter side of validation images to $256$ pixels and then crop the central region of $224 \times 224$ size for evaluation.

As shown in Table~\ref{tab:imagenet}, network models equipped with PSConv layers demonstrate consistent improvement over counterpart baseline models mostly with over 1\% gains of the top-1 error. We replace all the $3 \times 3$ standard convolutional layers in the middle of bottleneck blocks with our PSConv layers. In all of our main experiments, the cyclic interval is set to 4 and the dilation rate pattern is fixed as $\{d_1, d_2, d_3, d_4\} = \{1, 2, 1, 4\}$ which are determined by ablation studies, as detailed in the next subsection. It is observed that the PS-ResNet-50 model achieves 21.126\% top-1 error, which is comparable to the vanilla ResNet-101 model with almost half of the trainable parameter storage and computational resource consumption. The PS-ResNeXt-50 (32x4d) model even achieves superior performance over the vanilla 101-layer ResNeXt model, which demonstrates the wide applicability of our PSConv in boosting both standard and group convolution. Furthermore, we integrate PSConv into the modern SE-ResNet models and obtain performance margins again, which showcases the compatibility of our proposed convolution operation to other advanced atomic operations such as the channel-attention modules. Notably, all the above gains are obtained without theoretically introducing any additional computational cost.
\begin{table}[htbp]
	\caption{Recognition error comparisons on the ImageNet validation set. The standard metrics of top-1/top-5 errors are measured using single center crop evaluation. The baseline results are re-implemented by ourselves.}
	\label{tab:imagenet}
	\vskip -0.1in
	\centering
	\resizebox{.45\linewidth}{!}{
		\begin{tabular}{c|c|c}
			\toprule[0.2em]
			Architecture & Conv Type & Top-1 / Top-5 Err.(\%) \\
			\midrule[0.2em]
			\multirow{2}*{ResNet-50} 
			& standard & 22.850 / 6.532 \\
			& PSConv & \bf{21.126 / 5.724} \\
			\hline
			\multirow{2}*{ResNeXt-50 (32x4d)}
			& standard & 21.802 / 6.084 \\
			& PSConv & \bf{20.378 / 5.296} \\
			\hline
			\multirow{2}*{SE-ResNet-50}
			& standard & 22.192 / 6.040 \\
			& PSConv & \bf{20.814 / 5.578} \\
			\bottomrule[0.2em]
		\end{tabular}
	}
	\resizebox{.45\linewidth}{!}{
		\begin{tabular}{c|c|c}
			\toprule[0.2em]
			Architecture & Conv Type & Top-1 / Top-5 Err.(\%) \\
			\midrule[0.2em]
			\multirow{2}*{ResNet-101}
			& standard & 21.102 / 5.696 \\
			& PSConv & \bf{19.954 / 5.052} \\
			\hline
			\multirow{2}*{ResNeXt-101 (32x4d)}
			& standard & 20.502 / 5.390 \\
			& PSConv & \bf{19.498 / 4.724} \\
			\hline
			\multirow{2}*{SE-ResNet-101}
			& standard & 20.732 / 5.406 \\
			& PSConv & \bf{19.786 / 4.924} \\
			\bottomrule[0.2em]
		\end{tabular}
	}
	\vskip -0.1in
\end{table}
\newcommand{\tabincell}[2]{\begin{tabular}{@{}#1@{}}#2\end{tabular}}
\begin{table}[htbp]
	\caption{Performance comparison with state-of-the-art multi-scale network architectures on the ImageNet validation set.}
	\label{tab:comparison}
	\vskip -0.1in
	\centering
	\resizebox{.6\linewidth}{!}{
		\begin{tabular}{c|c|c|c|c}
			\toprule[0.2em]
			Network & Params & GFLOPs & LR decay schedule & Top-1 / Top-5 Err.(\%) \\
			\midrule[0.2em]
			ResNet-50~\cite{He_2016_CVPR} & 25.557M & 4.089 & cosine (120 epoch) & 22.850 / 6.532 \\
			\textbf{PS-ResNet-50 (ours)} & 25.557M & 4.089 & cosine (120 epoch) & \bf{21.126 / 5.724} \\
			\hline
			DRN-A-50~\cite{Yu_2017_CVPR} & 25.557M & 19.079 & stepwise (120 epoch) & 22.9 / 6.6 \\
			\hline
			DRN-D-54~\cite{Yu_2017_CVPR} & 35.809M & 28.487 & stepwise (120 epoch) & 21.2 / 5.9 \\ 
			\hline
			FishNet-150~\cite{NIPS2018_7356} & 24.96M & 6.45 & stepwise (100 epoch) & 21.86 / 6.05 \\
			\hline
			FishNet-150~\cite{NIPS2018_7356} & 24.96M & 6.45 & \tabincell{c}{cosine (200 epoch) \\ w/ label smoothing} & 20.65 / 5.25 \\ 
			\hline
			HRNet-W18-C~\cite{2019arXiv190404514S} & 21.3M & 3.99 & stepwise (100 epoch) & 23.2 / 6.6 \\ 
			\hline
			\tabincell{c}{OctResNet-50~\cite{chen2019drop}\\($\alpha=0.5$)} & 25.6M & 2.4 & cosine (110 epoch) & 22.6 / 6.4 \\
			\hline
			\tabincell{c}{bL-ResNet-50~\cite{chen2018biglittle}\\($\alpha=2,\beta=4$)} & 26.69M & 2.85 & cosine (110 epoch) & 22.69 / - \\
			\hline
			\tabincell{c}{Res2Net-50~\cite{gao2019res2net}\\($26w \times 4s$)} & 25.70M & 4.2 & stepwise (100 epoch) & 22.01 / 6.15 \\ 
			\hline
			ScaleNet-50~\cite{Li_2019_CVPR} & 31.483M & 3.818 & stepwise (100 epoch) & 22.02 / 6.05 \\ 
			\bottomrule[0.2em]
		\end{tabular}
	}
	\vskip -0.25in
\end{table}

For horizontal comparison, we give a brief synopsis of some recent multi-scale networks in Table~\ref{tab:comparison} for reference. Despite that discrepancies in model profiles and training strategies could lead to no apple-to-apple comparisons in most cases, our PS-ResNet-50 achieves competitive accuracy compared to other ResNet-50-based architectures under the similar level of parameters and computational complexities. Specifically, two variants of Dilated Residual Networks (DRN) increase the computation cost to a large extent due to the removed strides in the last two residual stages, but only achieves inferior or comparable performance.

\subsection{Ablation and Analysis}
\label{sec:ablation}
We first systematically probe the impact of partition numbers and dilation rate patterns in one cycle. We next assess the ability of PSConv to generalize to another classification benchmark beyond ImageNet, namely CIFAR-100.

\textbf{Partition Number.} On the one hand, provided that channels are divided into too many partitions, there leaves limited room for varied dilation rates within one partition and it frequently alternates around certain values. In the extreme case that the partition number equals to the number of channels, PSConv degenerates into the vanilla dilated convolution with a shared dilation rate. On the other hand, if there are too few partitions, each partition can accommodate a large number of heterogeneous dilation rates, which may have contradictory effects on extracting diverse features, hence we initially constrain the dilation rate in one basic pattern to toggle between 1 and 2 in this set of ablation experiments. Specifically, we set the dilation rate in one slot of a cycle to 2 and the other slots to 1. Under this constraint, features corresponding to large receptive fields will infrequently emerge with the growing cyclic interval, which may still impede the full utilization of multi-scale features. 
\begin{table}[htbp]
	\caption{Performance comparison of PS-ResNet-50 with varied cyclic intervals on the ImageNet validation set. The best result is highlighted in \textbf{bold}, the same hereinafter.}
	\label{tab:partition}
	\vskip -0.1in
	\centering
	\resizebox{.8\linewidth}{!}{
		\begin{tabular}{c|c|c|c|c}
			Architecture & ResNet-50 & \multicolumn{3}{c}{PS-ResNet-50} \\
			\midrule[0.1em]
			Cyclic Interval & $t=1$ (baseline) & $t=2$ & $t=4$ & $t=8$ \\
			Top-1/Top-5 Err.(\%) & \hspace{1.0em}22.850/6.532\hspace{1.0em} & \hspace{1.0em}21.948/5.978\hspace{1.0em} & \hspace{1.0em}\bf{21.476/5.720}\hspace{1.0em} & \hspace{1.0em}21.634/5.816\hspace{1.0em} \\
		\end{tabular}
	}
	\vskip -0.1in
\end{table}
\begin{table}[htbp]
	\caption{Performance comparison of PS-ResNet-50 with various dilation patterns on the ImageNet validation set.}
	\label{tab:pattern}
	\vskip -0.1in
	\centering
	\resizebox{\linewidth}{!}{
		\begin{tabular}{c|c|c|c|c|c}
			Dilation Pattern & $\{1, 1, 1, 1\}$ (\textbf{baseline}) & $\{1, 2, 1, 1\}$ & $\{1, 4, 1, 1\}$ & $\{1, 2, 1, 2\}$ & $\{1, 2, 1, 4\}$ (\textbf{default}) \\
			\midrule[0.1em]
			Top-1/Top-5 Err.(\%) & \hspace{1.0em}22.850/6.532\hspace{1.0em} & \hspace{1.0em}22.368/6.214\hspace{1.0em} & \hspace{1.0em}22.754/6.470\hspace{1.0em} & \hspace{1.0em}21.948/5.978\hspace{1.0em} & \hspace{1.0em}\bf{21.126/5.724}\hspace{1.0em} \\
		\end{tabular}
	}
	\vskip -0.3in
\end{table}

We use the ResNet-50 model on the ImageNet dataset for experiments and tune the partition numbers, giving rise to a spectrum of cyclic intervals. The corresponding results shown in Table~\ref{tab:partition} empirically support our speculation above. The PS-ResNet-50 ($t=4$) achieves better performance when the cyclic interval increases from 2 to 4. The accuracy tends to decline when its cyclic interval gets further increment. Thus we set $t=4$ as the default value in our main experiments. In each case, PS-ResNet-50 with a specific cyclic setting outperforms the vanilla ResNet-50 baseline result.

\textbf{Pattern of Dilation Rates.} Let the cyclic interval be 4. Noticing that the dilation rate pattern is an unordered set, we initially set any one of the dilation rate to a larger numeric value. For example, $\{d_1, d_2, d_3, d_4\}$ is set to $\{1, 2, 1, 1\}$, where the unique large dilation rate is placed in the second slot without loss of generality owing to its unordered nature. Next we assume that further increasing this large dilation rate (\text{e}.\textit{g}., setting $\{d_1, d_2, d_3, d_4\} = \{1, 4, 1, 1\}$) would lead to intra-group separation of these two dilation rates and unsmoothed transition of the receptive fields. Then we tend to inject another large dilation rate into this pattern. Considering that the setting of $\{d_1, d_2, d_3, d_4\} = \{1, 2, 1, 2\}$ is equivalent to $t=2$ in the above experiments, we change the pattern to $\{d_1, d_2, d_3, d_4\} = \{1, 2, 1, 4\}$ for the sake of perceiving larger receptive fields and interspacing the two different large dilation rates. This consequent PS-ResNet-50 achieves 21.126\% top-1 error in the ImageNet evaluation, which is exactly the one reported in Table~\ref{tab:imagenet}. For further exploration, we tentatively incorporate larger dilation rate to compose the combination of $\{d_1, d_2, d_3, d_4\} = \{1, 2, 4, 8\}$, but it shows much inferior performance (over 5\% drop). We attribute this failure to the exclusively aggressive dilation rate arrangement, since inappropriately enlarging the receptive field can involve irrelevant pixels into spatial correlation.

Apart from the static setting of dilation rates, we develop a learnable binary mask to distinguish the large dilation rate from the small one. This binary mask is decomposed via the Kronecker product, where the STE (Straight-Through Estimator)~\cite{10.1007/978-3-319-46493-0_32} technique is utilized to solve the discrete optimization problem. As a consequence, the dynamic version of PS-ResNet-50 with optional dilation rates of 1 and 2 reduces the top-1 error to 21.138\%, that is close to the best-performing static PS-ResNet-50 ($t=4$, $\{d_1, d_2, d_3, d_4\} = \{1, 2, 1, 4\}$) involving larger dilation rates in its PSConv pattern. Although extra parameters and computational complexity result in no fair comparison, it opens up a promising perspective deserving future research development.
\begin{table}[htbp]
	\caption{Performance comparison of PS-ResNet-50 on the ImageNet validation set, with the variation of dilation rates along different axes of kernel lattice.}
	\label{tab:variation}
	\vskip -0.1in
	\centering
	\resizebox{.6\linewidth}{!}{
		\begin{tabular}{c|c|c}
			\toprule[0.2em]
			\hspace{1.0em}Input Channel Axis\hspace{1.0em} & \hspace{1.0em}Output Channel Axis\hspace{1.0em} & Top-1/Top-5 Err.(\%) \\
			\midrule[0.2em]
			\ding{51} & \ding{55} & 21.658/5.832\\
			\ding{55} & \ding{51} & 22.056/6.174\\
			\ding{51} & \ding{51} & \bf{21.126/5.724} \\
			\bottomrule[0.2em]
		\end{tabular}
	}
	\vskip -0.15in
\end{table}
\begin{table}[htbp]
	\caption{Top-1 Error comparisons on the CIFAR-100 test set. Our results were obtained by computing mean and standard deviation over 5 individual runs (denoted by mean $\pm$ std. in the table).}
	\label{tab:cifar}
	\vskip -0.1in
	\centering
	\resizebox{.49\linewidth}{!}{
		\begin{tabular}{c|c|c}
			\toprule[0.2em]
			Architecture & Conv Type & Top-1 Error(\%) \\
			\midrule[0.2em]
			\multirow{3}*{ResNeXt-29 (8x64d)} 
			& standard (\textit{official})\quad\quad & $17.77$ \\
			& standard (\textit{self impl.}) & $18.074 \pm 0.130$ \\
			& PSConv & $\bf{17.138 \pm 0.286}$ \\
			\bottomrule[0.2em]
		\end{tabular}
	}	
	\resizebox{.49\linewidth}{!}{
		\begin{tabular}{c|c|c}
			\toprule[0.2em]
			Architecture & Conv Type & Top-1 Error(\%) \\
			\midrule[0.2em]
			\multirow{3}*{ResNeXt-29 (16x64d)}
			& standard (\textit{official})\quad\quad & $17.31$ \\
			& standard (\textit{self impl.}) & $17.538 \pm 0.094$ \\
			& PSConv & $\bf{16.528 \pm 0.353}$ \\
			\bottomrule[0.2em]
		\end{tabular}
	}
	\vskip -0.25in
\end{table}

Following the searched optimal setting of $\{d_1, d_2, d_3, d_4\} = \{1, 2, 1, 4\}$, we remove the shift strategy among different filters, which means the variation of dilation rates only exists in the axis of input channels. In this setup, we observe a drop of around 1\% regarding the top-1 validation accuracy. Symmetrically, we only vary the dilation rates in the axis of output channels with the same setting of $\{d_1, d_2, d_3, d_4\} = \{1, 2, 1, 4\}$, which indicates no cyclic operations inside each individual filter. As shown in Table~\ref{tab:variation}, it also achieves inferior performance.

\textbf{Beyond ImageNet.} CIFAR-100~\cite{Krizhevsky09} is another widely-adopted benchmark for image classification, which consists of 50,000 training images and 10,000 test images. Each colorful image in the dataset is of $32 \times 32$ size and drawn from 100 classes, hence it is more challenging than CIFAR-10 with similar image qualities but a coarser taxonomy. We choose the high-performing ResNeXt~\cite{Xie_2017_CVPR} architecture as a strong baseline, and replace all the $3 \times 3$ convolutional layers in every bottleneck block with PSConv layers to build our PS-ResNeXt models for comparison. The data augmentation is the same as the preprocessing method in~\cite{He_2016_CVPR,Xie_2017_CVPR}, utilizing sequential zero padding, random cropping and standardization. The whole training regime strictly follows the original paper to isolate the contribution of our PSConv. For evaluation, we perform five independent runs of training the same architecture with different initialization seeds and report the mean top-1 error as well as the standard deviation.

We summarize the comparison results in Table~\ref{tab:cifar}. The performance of our reproduced ResNeXt-29 is slightly degraded, thus we list results from both the official release and our implementation, annotated as \textit{official} and \textit{self impl.} with the standard convolution respectively. It is evident that PS-ResNeXt-29 (8x64d) and PS-ResNeXt-29 (16x64d) outperform the original ResNeXts by around 1\% accuracy gains. Even compared to the results from the original author, absolute gains of 0.632\% and 0.782\% are achieved using our PSConv neural networks. It is observed that using networks with various cardinalities on datasets with distinct characteristics (like thumbnails), PSConv could still yield satisfactory performance gains.

\textbf{Speed Benchmark.} For an input tensor with the size of $(N, C, H, W)=(200, 64, 56, 56)$, a standard $3 \times 3$ convolutional layer with 64 output channels takes 4.85ms to process on a single Titan X GPU, using CUDA v9.0 and cuDNN v7.0 as the backend. The dilated convolution with a dilation rate of 2 consumes 2.99 times of above and the inference time of our PSConv is 1.14$\times$ of dilated convolution. There exist a similar trend in the comparison of their group convolution based counterparts. Thus, improved performance of inference speed can be achieved by optimizing vanilla dilated convolutions on GPU/CPU inference. The further optimized results for practical deployment are provided in the supplementary materials.

\textbf{Scale Allocation.} We dive into the PSConv kernels to analyze the law of scale-relevant feature distributions by dissecting the weight proportion with respect to different dilation rates, as is shown in the supplementary materials.

\subsection{MS COCO 2017}

To further demonstrate the generality of our proposed convolution, we apply the PSConv-based backbones to object detection and instance segmentation frameworks and finetune the PSConv-based detectors on the 2017 version of Microsoft COCO~\cite{10.1007/978-3-319-10602-1_48} benchmark. This large-scale dataset including 118,287 training images and 5,000 validation images is considered highly challenging owing to the huge number of objects within per image and large variation among these instances, which is suitable for inspecting the superiority of our PSConv models.

We use the popular MMDetection~\cite{2019arXiv190607155C} toolbox to conduct experiments. ResNet-50/101 and ResNeXt-101 (32x4d) along with FPN~\cite{Lin_2017_CVPR} necks are selected as the backbone networks. For object detection and instance segmentation tasks, we adopt the main-stream Faster R-CNN~\cite{NIPS2015_5638} and Mask R-CNN~\cite{He_2017_ICCV} as the basic detectors respectively. We replace all the $3 \times 3$ convolutional layers in the pre-trained backbone network by PSConv layers, while the convolution layers in the FPN neck are kept as standard convolutions\footnote{Actually we have preliminary experiments by also replacing these layers with PSConv layers, but it achieves marginal benefit. For instance, AP\textsuperscript{bbox} of Faster R-CNN with ResNet-50 and FPN only increases from 38.4 to 38.6.}. Then we finetune these detectors on the training set following the $1\times$ learning rate schedule, which indicates a total of 12 epochs with the learning rate divided by 10 at the epoch of 8\textsuperscript{th} and 11\textsuperscript{st} respectively. During this transfer learning process, we maintain the same data preparation pipeline and hyperparameter settings for our models as the baseline detectors. During evaluation, we test on the validation set and report the COCO-style Average Precision (AP) under IOU thresholds ranging from 0.5 to 0.95 with an increment of 0.05. We also keep track of scores for small, medium and large objects. These metrics comprehensively assess the qualities of detection and segmentation results from various views of different scales.

\begin{table}[htbp]
	\caption{Bounding-box and mask Average Precision (AP) comparison on the COCO 2017 validation set for the instance segmentation track with different backbones.}
	\label{tab:mrcnn}
	\vskip -0.1in
	\centering
	\resizebox{\linewidth}{!}{
		\begin{tabular}{c|c|c|c|c|c|c|c|c|c|c|c|c|c|c}
			\toprule[0.2em]
			\multirow{2}*{Detector} & \multirow{2}*{Architecture} & \multirow{2}*{Conv Type} & \multicolumn{6}{|c|}{Box AP} & \multicolumn{6}{|c}{Mask AP} \\
			\cline{4-15}
			& & & AP & $\text{AP}_{50}$ & $\text{AP}_{75}$ & $\text{AP}_{S}$ & $\text{AP}_{M}$ & $\text{AP}_{L}$ & AP & $\text{AP}_{50}$ & $\text{AP}_{75}$ & $\text{AP}_{S}$ & $\text{AP}_{M}$ & $\text{AP}_{L}$ \\
			\midrule[0.2em]
			\multirow{6}*{Mask R-CNN} 
			& \multirow{2}*{R50} & standard & 37.3\hspace{2.5em} & 59.0 & 40.2 & 21.9 & 40.9 & 48.1 & 34.2\hspace{2.5em} & 55.9 & 36.2 & 15.8 & 36.9 & 50.1 \\
			& & PSConv & 39.4\textsubscript{(\textbf{+2.1})} & 61.3 & 42.8 & 24.1 & 43.1 & 51.3 & 35.6\textsubscript{(\textbf{+1.4})} & 57.9 & 37.9 & 17.2 & 38.4 & 52.4 \\
			\cline{2-15}
			& \multirow{2}*{R101} & standard & 39.4\hspace{2.5em} & 60.9 & 43.3 & 23.0 & 43.7 & 51.4 & 35.9\hspace{2.5em} & 57.7 & 38.4 & 16.8 & 39.1 & 53.6 \\
			& & PSConv & 41.6\textsubscript{(\textbf{+2.2})} & 63.4 & 45.1 & 24.7 & 45.6 & 54.4 & 37.4\textsubscript{(\textbf{+1.5})} & 60.0 & 39.8 & 17.8 & 40.4 & 55.1 \\
			\cline{2-15}
			& \multirow{2}*{X101-32x4d} & standard & 41.1\hspace{2.5em} & 62.8 & 45.0 & 24.0 & 45.4 & 52.6 & 37.1\hspace{2.5em} & 59.4 & 39.7 & 17.7 & 40.5 & 53.8 \\
			& & PSConv & 42.4\textsubscript{(\textbf{+1.3})} & 64.4 & 46.1 & 25.4 & 46.5 & 55.7 & 38.0\textsubscript{(\textbf{+0.9})} & 60.8 & 40.5 & 18.6 & 41.0 & 55.8 \\
			\hline
			\multirow{6}*{Cascade Mask R-CNN} 
			& \multirow{2}*{R50} & standard & 41.2\hspace{2.5em} & 59.1 & 45.1 & 23.3 & 44.5 & 54.5 & 35.7\hspace{2.5em} & 56.3 & 38.6 & 16.4 & 38.2 & 52.6 \\
			& & PSConv & 42.9\textsubscript{(\textbf{+1.7})} & 61.7 & 46.9 & 24.2 & 46.5 & 57.2 & 36.9\textsubscript{(\textbf{+1.2})} & 58.4 & 39.5 & 17.1 & 39.4 & 54.6 \\
			\cline{2-15}
			& \multirow{2}*{R101} & standard & 42.6\hspace{2.5em} & 60.7 & 46.7 & 23.8 & 46.4 & 56.9 & 37.0\hspace{2.5em} & 58.0 & 39.9 & 16.7 & 40.3 & 54.6 \\
			& & PSConv & 44.6\textsubscript{(\textbf{+2.0})} & 63.2 & 48.6 & 25.9 & 48.7 & 59.6 & 38.4\textsubscript{(\textbf{+1.4})} & 60.5 & 41.2 & 18.6 & 41.5 & 56.8 \\
			\cline{2-15}
			& \multirow{2}*{X101-32x4d} & standard & 44.4\hspace{2.5em} & 62.6 & 48.6 & 25.4 & 48.1 & 58.7 & 38.2\hspace{2.5em} & 59.6 & 41.2 & 18.3 & 41.4 & 55.6 \\
			& & PSConv & 45.3\textsubscript{(\textbf{+0.9})} & 64.2 & 49.5 & 27.0 & 49.2 & 60.0 & 38.9\textsubscript{(\textbf{+0.7})} & 61.1 & 41.8 & 19.4 & 41.9 & 56.6 \\
			\bottomrule[0.2em]
		\end{tabular}
	}
	\vskip -0.2in
\end{table}

The comparison results of Mask R-CNN are shown in Table~\ref{tab:mrcnn} (similar comparisons of Faster R-CNN are provided in the supplementary materials), where the baseline results with standard backbone networks are extracted from the model zoo of MMDetection, and absolute gains of AP concerning our PSConv models are indicated in the parentheses. Since our ImageNet pre-trained backbones in Section~\ref{sec:ilsvrc12} are trained using the cosine learning rate annealing, we would have an unfair accuracy advantage against the pre-trained backbones in MMDetection. In order to pursue fair comparison to its published baseline results, we first retrain backbones of ResNet and ResNeXt following the conventional step-wise learning rate annealing strategy~\cite{He_2016_CVPR} and then load these backbones to the detectors\footnote{If we adopt those unfair backbones pre-trained using cosine learning rate decay in Section~\ref{sec:ilsvrc12}, we can get even larger performance margins (\textit{e}.\textit{g}. 2.6\% instead of 2.0\% for Faster R-CNN with ResNet-50 and FPN).}. It is evident that PSConv brings consistent and considerable performance gains over the baseline results, across different tasks and various backbones. In addition, we introduce the Cascade (Mask) R-CNN~\cite{Cai_2018_CVPR} as a stronger baseline detector and reach the conclusion that our PSConv operation can benefit both basic detectors and more advanced cascade detectors.

Detectors with the ResNet-101 backbone consistently show larger margins among different tasks and frameworks compared to the other two backbones. Compared to ResNet-50, the 101-layer network almost quadruples the depth of the conv4\_x stage, guaranteeing a higher capacity for performance amelioration. In addition, we come up with the hypothesis that its ResNeXt counterpart has already efficiently deployed the model capacity through adjusting the dimension of cardinality beyond network depth and width, leaving a bottleneck for further performance improvement in both classification and detection tasks. It is observed that the most significant improvement of Faster R-CNN and Mask R-CNN locates in the metric of $\text{AP}_{L}$ among various object sizes, speaking to the theoretically enlarged receptive fields. Finally, representative visualization results of predicted bounding-boxes and masks are attached in the supplementary materials to raise the qualitative insight of our method.

\section{Conclusion}

In this paper, we have proposed a novel convolution operation named PSConv, which cyclically alternates dilation rates along the axes of input and output channels. PSConv permits to aggregate multi-scale features from a granular perspective and efficiently allocates weights to a collection of scale-specific features through dynamic execution. It is amenable to be plugged into arbitrary state-of-the-art CNN architectures in-place, demonstrating its superior performance on various vision tasks compared to the counterparts with regular convolutions.

\section*{\LARGE Appendix}
\appendix

\section{PSConv Based on Depthwise Convolution}

\begin{figure}[htbp]
	\begin{center}
		\includegraphics[width=.75\linewidth,trim=10 20 10 50,clip]{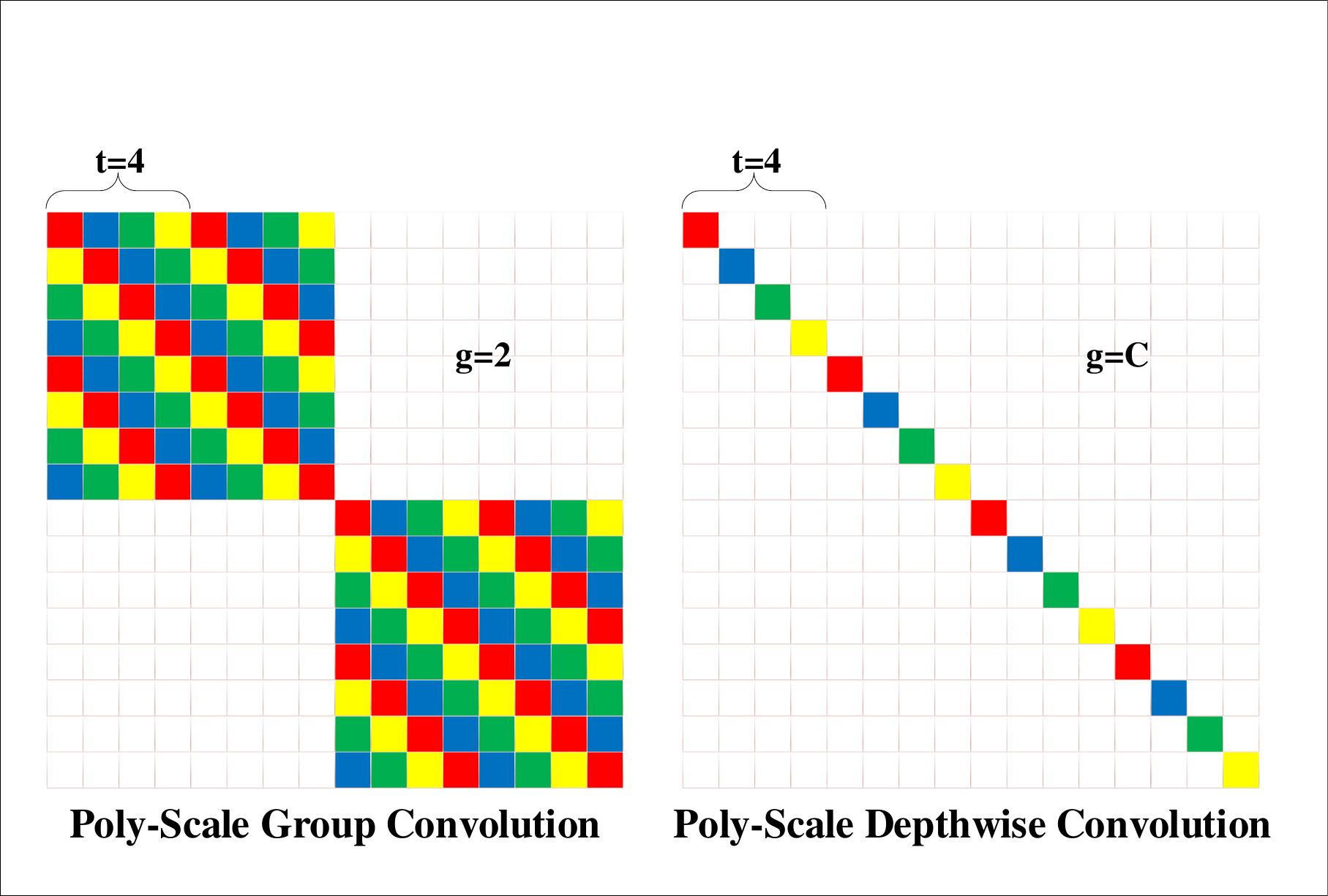}
	\end{center}
	\caption{Comparison between dilation space of kernel lattice in Poly-Scale group convolution (group number $g=2$ and cyclic interval $t=4$) and Poly-Scale depthwise convolution (group number $g=C$ and cyclic interval $t=4$), where $C=16$ represents the number of channels. Similar to Fig.~\ref{fig:conv} in the main paper, each color indicates one specific type of dilation rate, the same hereinafter. Best viewed in color.}
	\label{fig:dwconv}
\end{figure}

Regarding the variant of PSConv based on depthwise convolution (DWConv), we do not consider it in our main method because applying PSConv directly to DWConv is non-trivial. Each group of DWConv contains only one channel, thus the cyclic pattern cannot be accommodated inside one group. However, adapting our cyclic pattern to external groups is possible. Specifically, one pattern is arranged across $t$ groups, where $t$ is the original cyclic interval. The illustration diagram is depicted in Fig.~\ref{fig:dwconv}. It is also noted that such a DWConv-based variant is akin to the MixConv+dilated accompanied with channel shuffling.

\begin{figure}[htbp]
	\begin{center}
		\includegraphics[width=.75\linewidth,trim=10 20 10 50,clip]{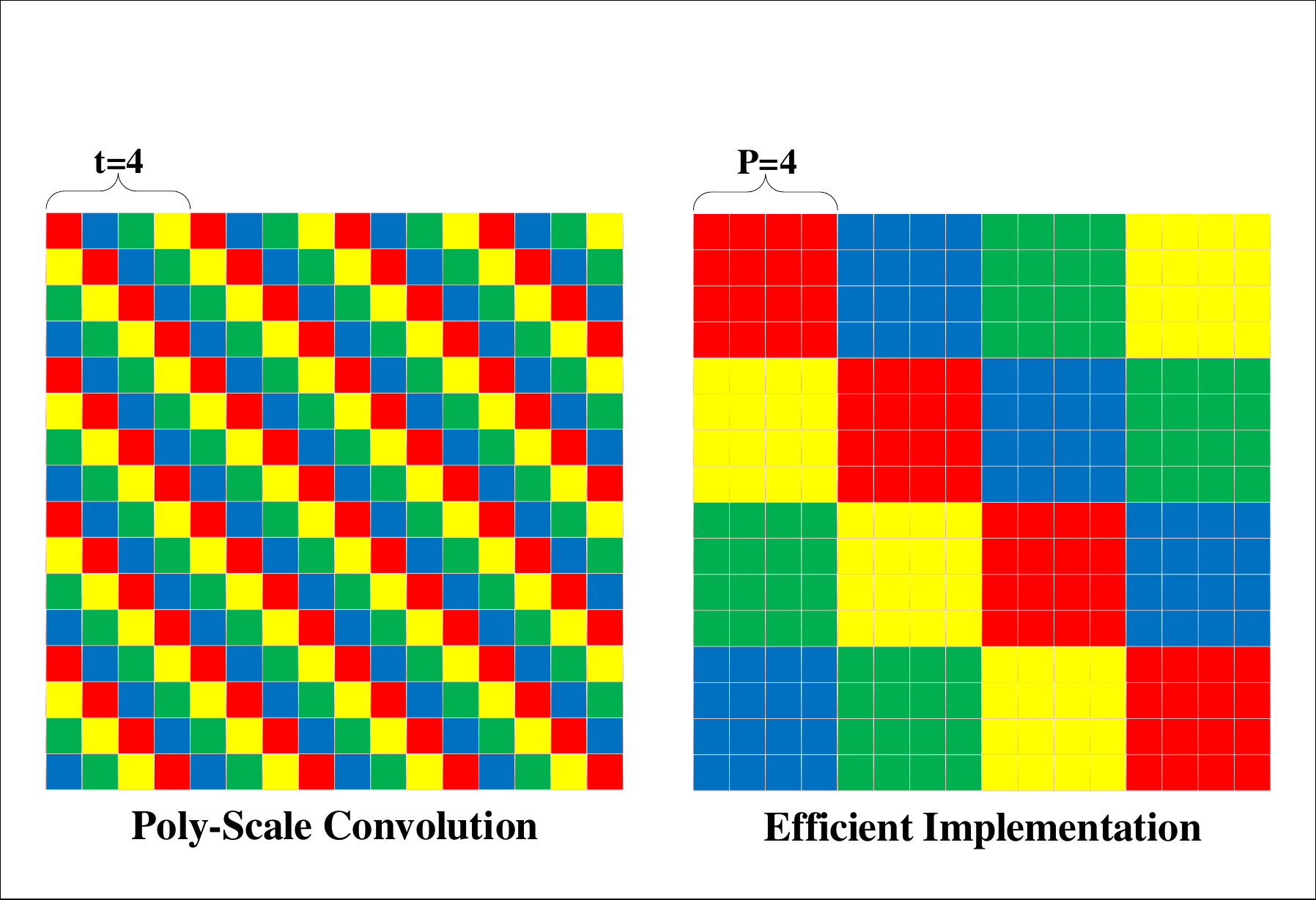}
	\end{center}
	\caption{Comparison between the dilation space of the original PSConv (cyclic interval $t=4$) and its rearranged dilation space for efficient implementation.}
	\label{fig:implementation}
\end{figure}

\begin{figure}[htbp]
	\begin{center}
		\includegraphics[width=.75\linewidth,trim=10 20 10 20,clip]{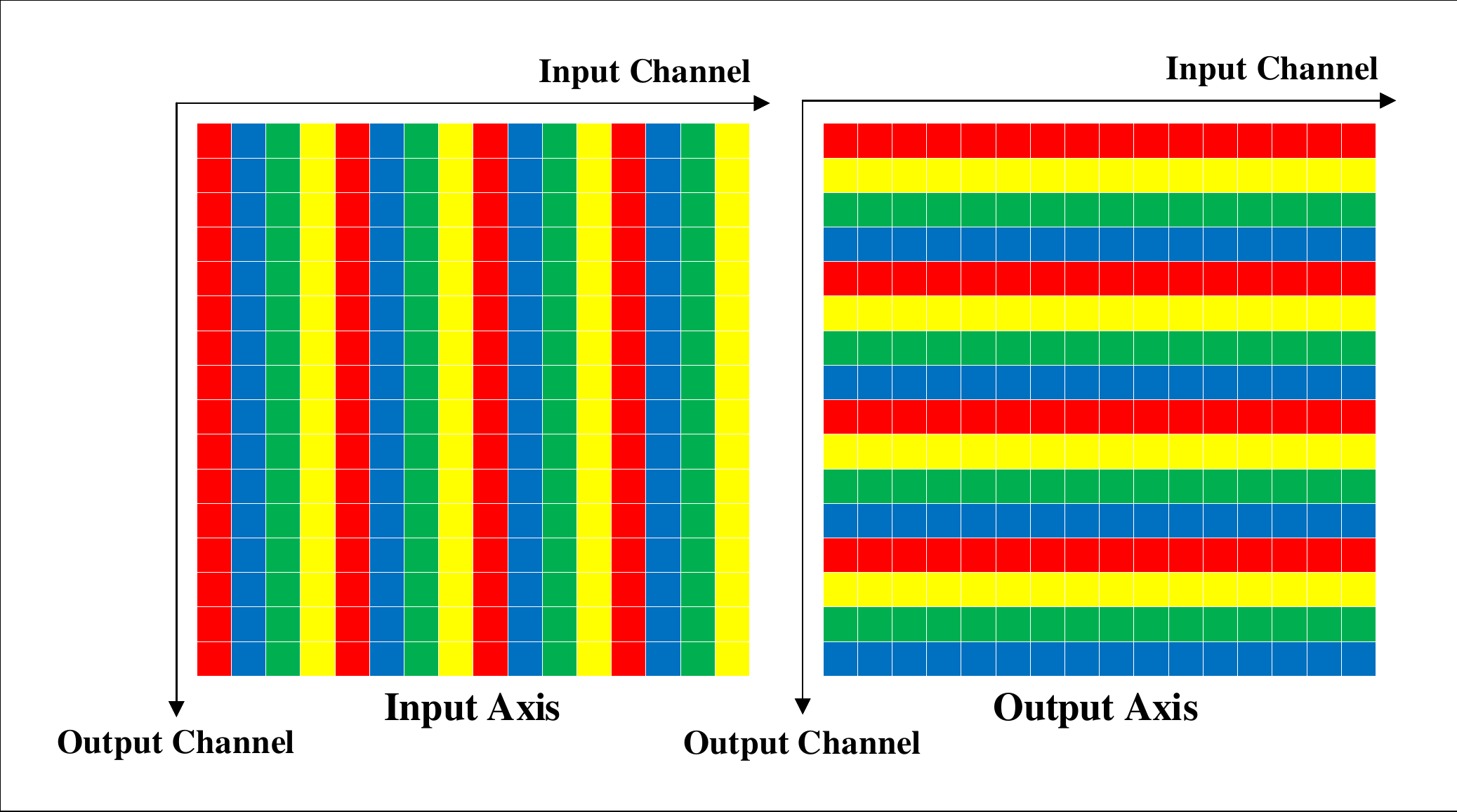}
	\end{center}
	\caption{The dilation space of two simplified cases of PSConv, which only vary dilation rates along the input (\textit{left}) or output (\textit{right}) channel axis.}
	\label{fig:ablation}
\end{figure}

\section{Efficient Implementation}

In view of the interchangeability of channel indices, we provide an equivalent but efficient implementation of the original PSConv by grouping channels with the same dilation rate together. For each row in the re-arranged dilation space of kernel lattice, the same dilation rates in each partition ($P$ partitions in total) are assembled, shaping a group with $P$ channels. When rearranging the input channel indices, the output channel indices are rearranged accordingly, since the input channels of the current layer are output channels of the precedent layer. The original and rearranged PSConv are comparatively illustrated in Fig.~\ref{fig:implementation}. Reminiscent of the definition in the main paper, the dilation rate matrix $D$ is a block matrix after rearrangement, which serves the purpose of efficient matrix operations.

\section{Ablation of Dilation Patterns}

To validate the effectiveness of our design principle, we develop two simplified cases for ablation studies, as shown in Fig.~\ref{fig:ablation}. The first one merely varies dilation rates along the input channel axis, which means removing the shift operation from PSConv. Actually it can be interpreted as splitting the incoming features into groups along the channel dimension, transforming these features with one dilation rate per group and aggregating the output features through summation. The second one merely varies dilation rates along the output channel axis. It can be interpreted as transforming the incoming features with different dilation rates in parallel and concatenating the output features along the channel dimension. Therefore, both of these two cases can reduce to the multi-scale network design from the filter space. In contrast, the original PSConv is a more granular design in the kernel space. The corresponding ablative experiments are discussed in the Section~\ref{sec:ablation} of the main paper and the comparison in Table~\ref{tab:variation} of the main paper also demonstrates the superiority of the original PSConv compared to these two simplified design.

\section{Visualization of Scale Allocation}

\begin{figure}[htbp]
	\begin{center}
		\includegraphics[width=.4\linewidth,trim=30 30 30 30,clip]{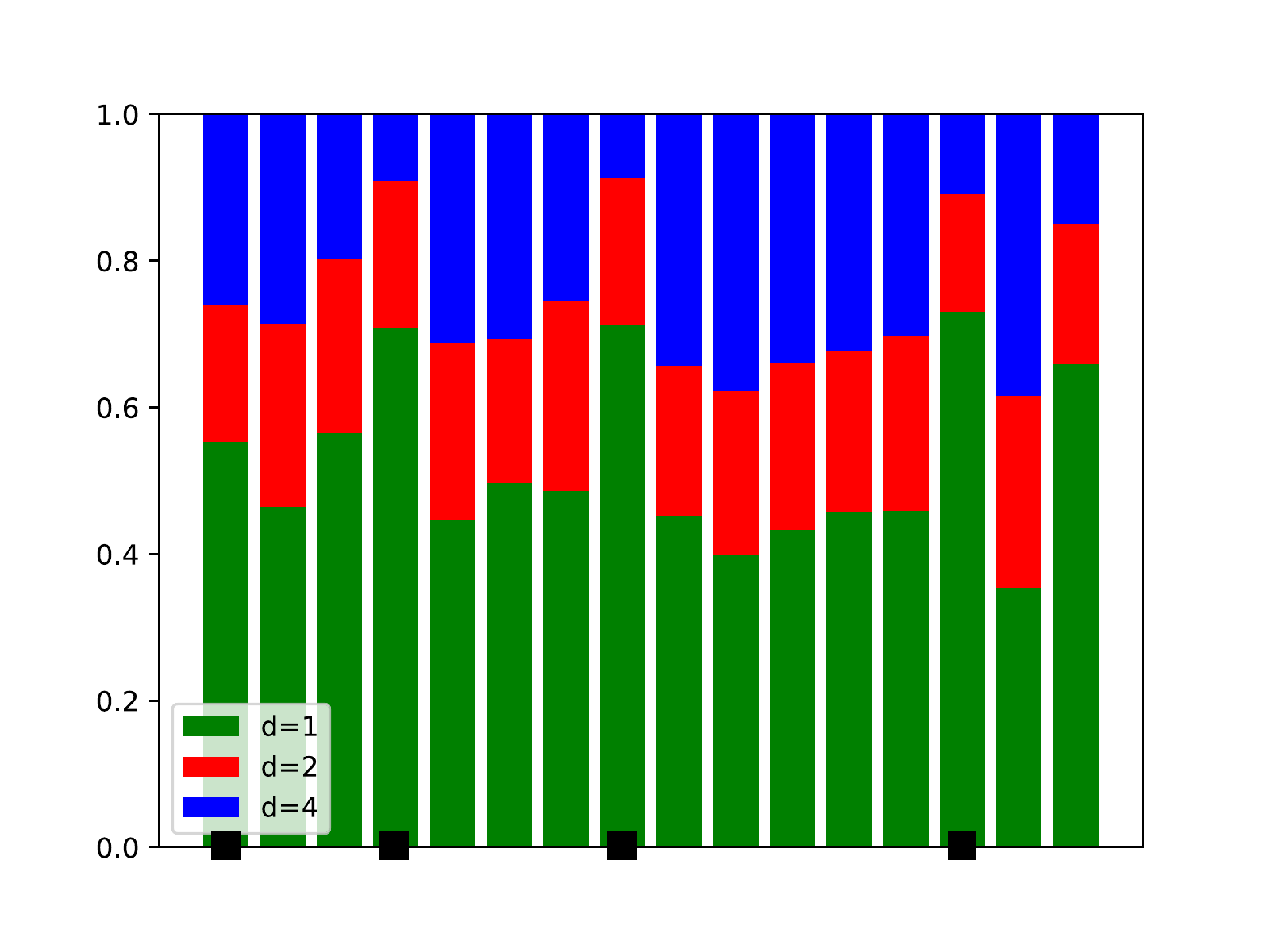}
		\includegraphics[width=.4\linewidth,trim=30 30 30 30,clip]{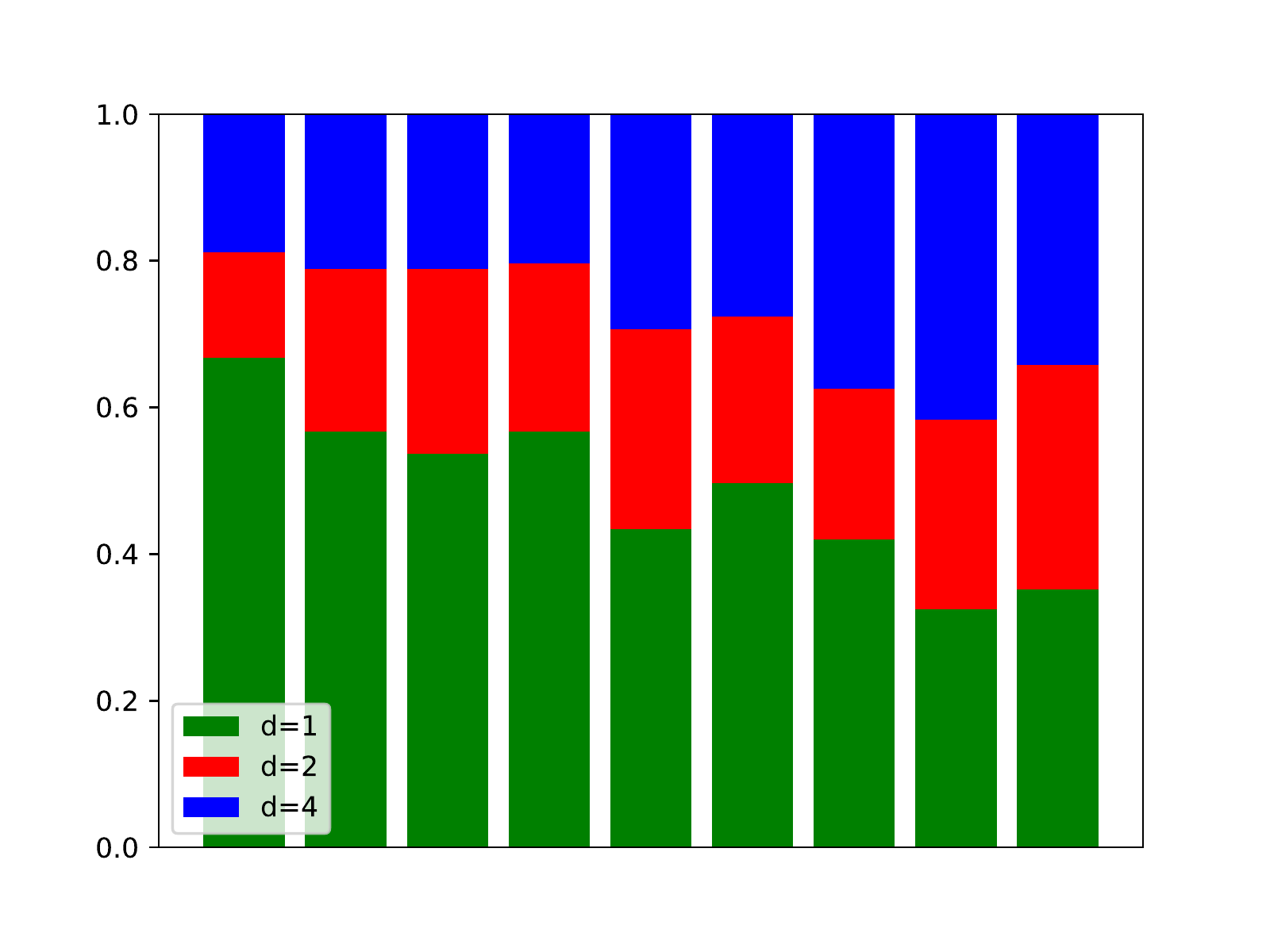}
	\end{center}
	\vskip -0.2in
	\caption{Visualization of the automated selection mechanism concerning multi-scale features. The left panel reveals the result of PS-ResNet-50 on the ImageNet, where the horizontal axis corresponds to indices of residual blocks, $\blacksquare$ indicates the starting block of stage 2-5. The right panel displays the result of PS-ResNeXt-29 (16$\times$64d) on the CIFAR-100. Best viewed in color.}
	\label{fig:viz}
\end{figure}

With curiosity about the learned distribution of scale-relevant features, we dissect the weight proportions with respect to different dilation rates in each PSConv layer, as illustrated in Fig.~\ref{fig:viz}. For each dilation rate in a PSConv layer, we compute the mean of absolute values in each $3 \times 3 $ kernel and take the maximum across all corresponding kernels as the proxy. These layer-wise proxies can be representative of the importance of different dilation rates. They are finally normalized inside each layer for inter-layer comparison.

As for PS-ResNet-50 on the ImageNet, it is observed that in the first residual block of stage 3-5 (\texttt{conv3\_x}, \texttt{conv4\_x}, and \texttt{conv5\_x}), where feature maps are processed with stride 2, PSConv is prone to overlook convolutional kernels with large dilation rates and emphasize those without dilation rates, as the downsampling operation already offers sufficient amplification of the receptive fields at these points. This trend is not obvious for PS-ResNeXt-29 (16$\times$64d) on CIFAR-100, partially due to its few down-sampling operations. Nevertheless, there exists a clear tendency that convolutional kernels with large dilation rates will occupy a larger proportion in the deeper layers, implying the necessity of allocating more resources to semantic features in the high-level layers. The visual analysis also helps understand the quantitative performance improvement with a better coarse-to-fine feature generation process compared to standard convolutions.

\section{Object Detection}

We perform experiments with Faster R-CNN on the MS COCO object detection track and report the results in Table~\ref{tab:frcnn}. Compared to the detectors with vanilla convolutions, PSConv also achieves obviously higher AP based on different backbone architectures. The comparison of performance gains across three backbone networks shows a similar trend as Mask R-CNN in the main paper.

\begin{table}[htbp]
	\caption{Bounding-box Average Precision (AP) comparison on the COCO 2017 validation set for the bounding-box detection track with different backbones.}
	\vskip -0.1in
	\label{tab:frcnn}
	\centering
	\resizebox{.8\linewidth}{!}{
		\begin{tabular}{c|c|c|c|c|c|c|c|c}
			\toprule[0.2em]
			Detector & Architecture & Conv Type & AP & $\text{AP}_{50}$ & $\text{AP}_{75}$ & $\text{AP}_{S}$ & $\text{AP}_{M}$ & $\text{AP}_{L}$ \\
			\midrule[0.2em]
			\multirow{6}*{Faster R-CNN} 
			& \multirow{2}*{R50} & standard & 36.4\hspace{2.5em} & 58.4 & 39.1 & 21.5 & 40.0 & 46.6 \\
			& & PSConv & 38.4\textsubscript{(\textbf{+2.0})} & 60.6 & 41.6 & 22.9 & 42.4 & 49.9 \\
			\cline{2-9}
			& \multirow{2}*{R101} & standard & 38.5\hspace{2.5em} & 60.3 & 41.6 & 22.3 & 43.0 & 49.8 \\
			& & PSConv & 40.9\textsubscript{(\textbf{+2.4})} & 63.0 & 44.3 & 23.8 & 45.3 & 53.5 \\
			\cline{2-9}
			& \multirow{2}*{X101-32x4d} & standard & 40.1\hspace{2.5em} & 62.0 & 43.8 & 23.4 & 44.6 & 51.7 \\
			& & PSConv & 41.3\textsubscript{(\textbf{+1.2})} & 63.6 & 45.1 & 24.7 & 45.5 & 53.8 \\
			\hline
			\multirow{6}*{Cascade R-CNN} 
			& \multirow{2}*{R50} & standard & 40.4\hspace{2.5em} & 58.5 & 43.9 & 21.5 & 43.7 & 53.8 \\
			& & PSConv & 41.9\textsubscript{(\textbf{+1.5})} & 60.8 & 45.5 & 24.2 & 45.3 & 55.6 \\
			\cline{2-9}
			& \multirow{2}*{R101} & standard & 42.0\hspace{2.5em} & 60.3 & 45.9 & 23.2 & 45.9 & 56.3 \\
			& & PSConv & 43.8\textsubscript{(\textbf{+1.8})} & 62.6 & 47.7 & 25.6 & 47.5 & 57.9 \\
			\cline{2-9}
			& \multirow{2}*{X101-32x4d} & standard & 43.6\hspace{2.5em} & 62.2 & 47.4 & 25.0 & 47.7 & 57.4 \\
			& & PSConv & 44.4\textsubscript{(\textbf{+0.8})} & 63.6 & 48.4 & 26.6 & 48.3 & 59.2 \\
			\bottomrule[0.2em]
		\end{tabular}
	}
\end{table}

\section{Visualization of Predictions on MS COCO}

We select Faster R-CNN and Mask R-CNN with ResNet-101 for visualization in view of the large margins between our PSConv-based detectors and the standard ones under this setting, as indicated by experimental results in the Section 4.3 of the main paper.

\begin{figure}[htbp]
	\begin{center}
		\includegraphics[width=.83\linewidth]{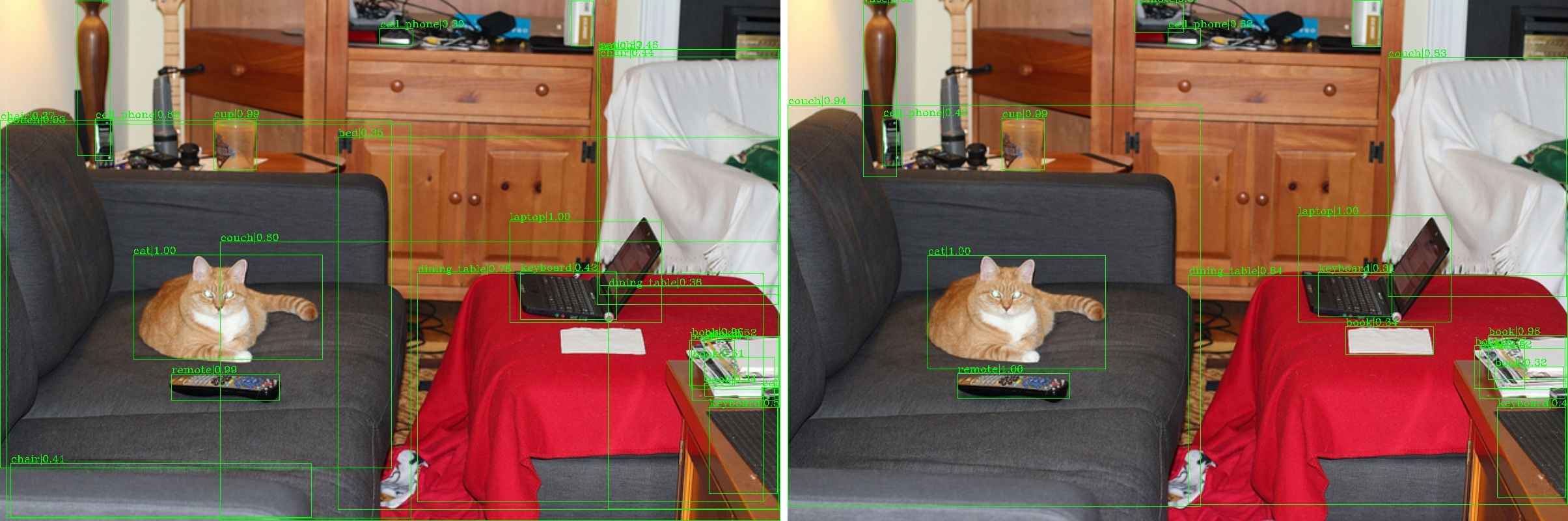}
		\includegraphics[width=.83\linewidth]{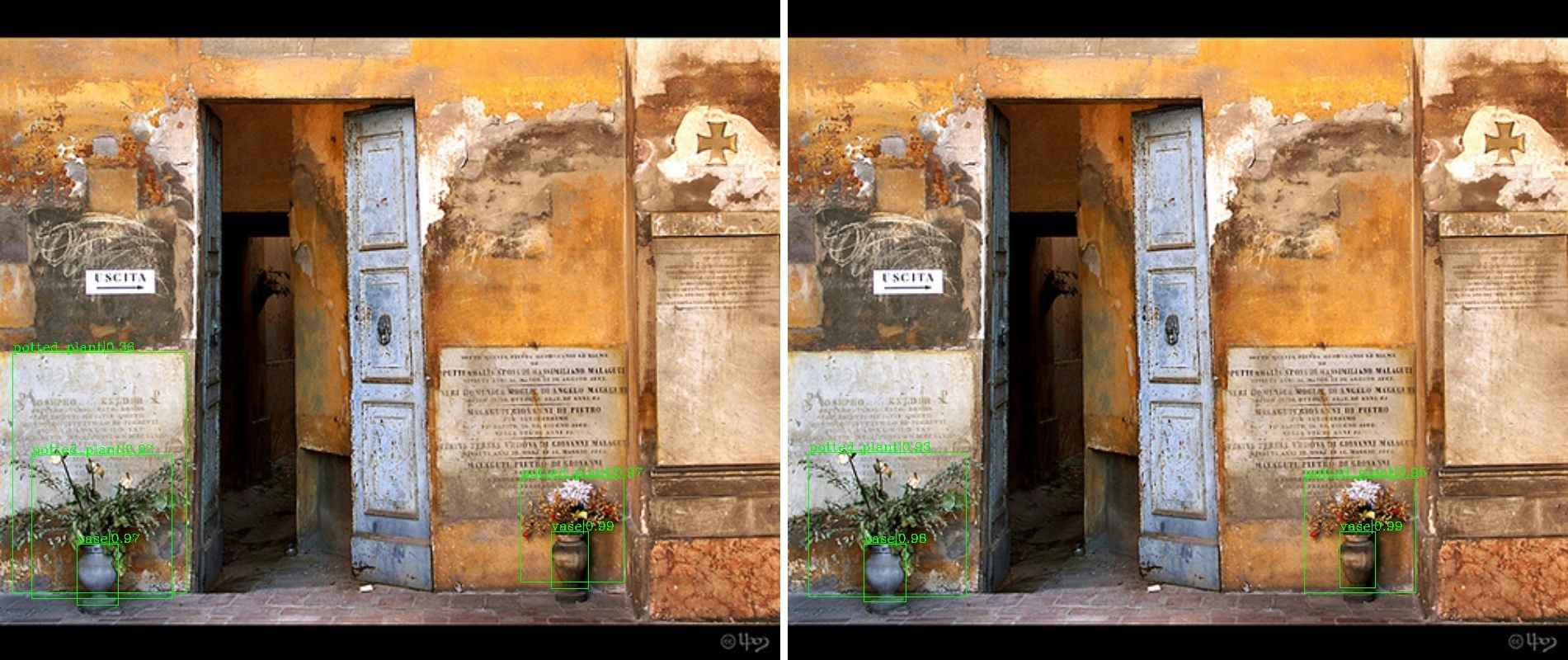}
		\includegraphics[width=.83\linewidth]{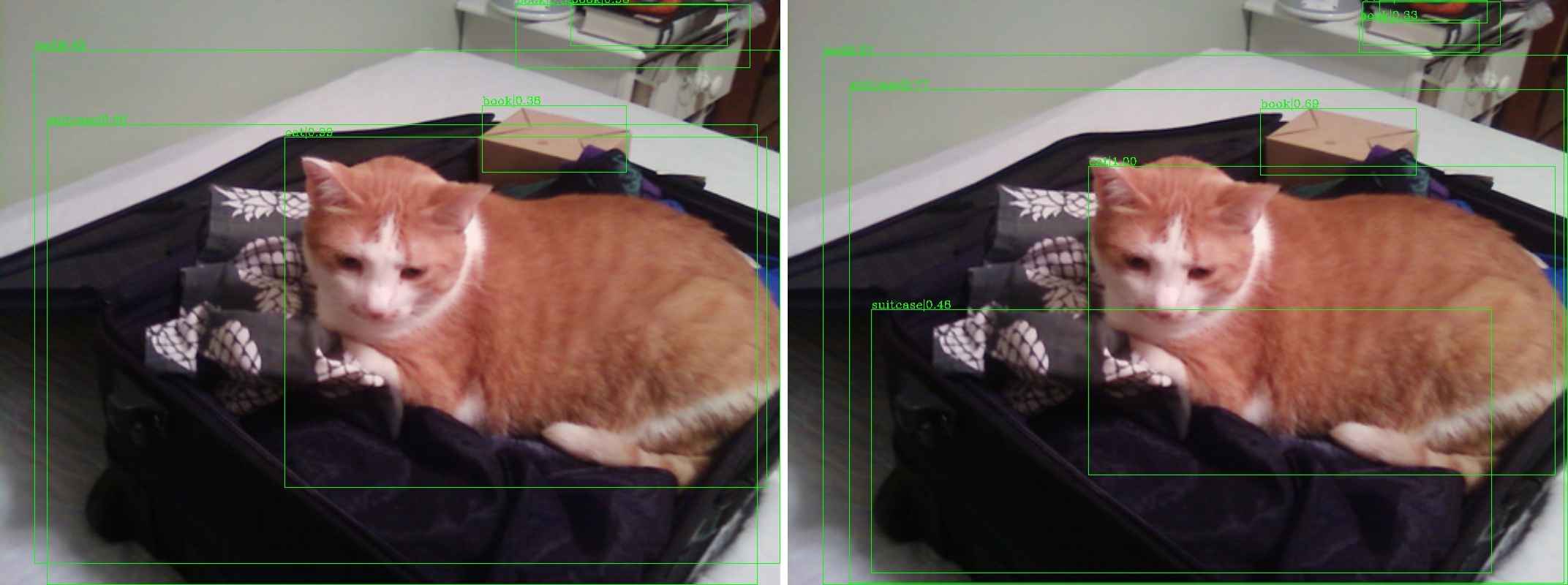}
		\includegraphics[width=.83\linewidth]{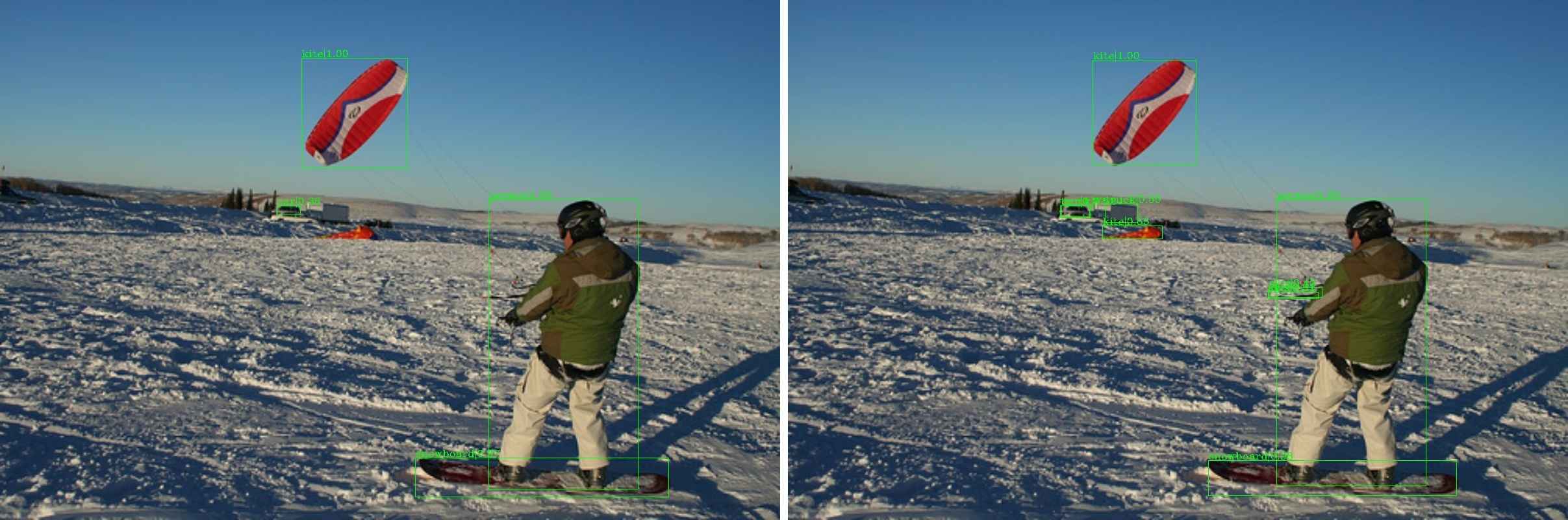}
	\end{center}
	\caption{Some bounding-box detection results of Faster R-CNN with ResNet-101 on the COCO 2017 validation set. The left panel shows predictions from the standard detector while the right panel shows the detector equipped with our PSConv, the same hereinafter.}
	\label{fig:frcnn-1}
\end{figure}

\begin{figure}[htbp]
	\begin{center}
		\includegraphics[width=.68\linewidth]{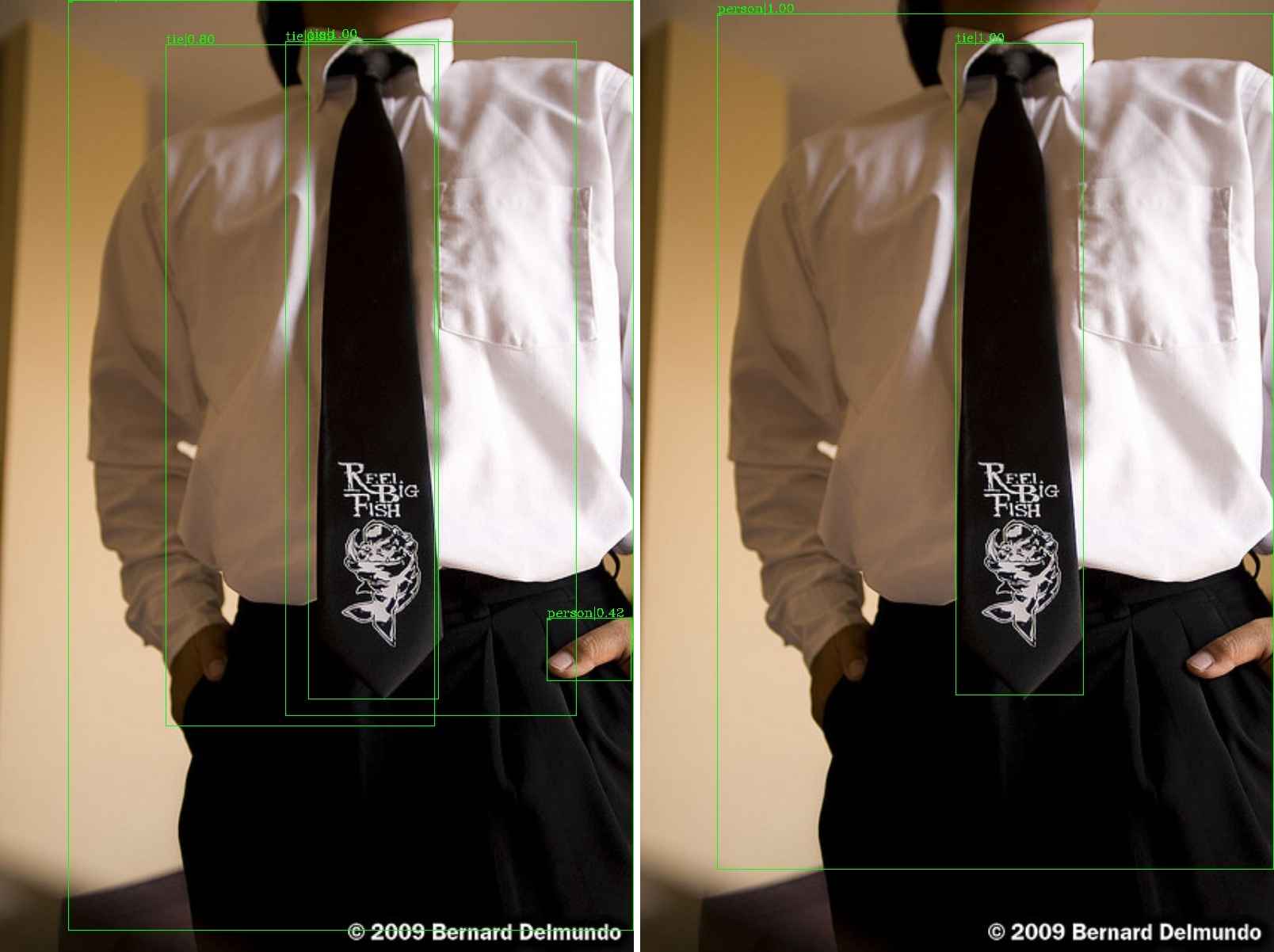}
		\includegraphics[width=.68\linewidth]{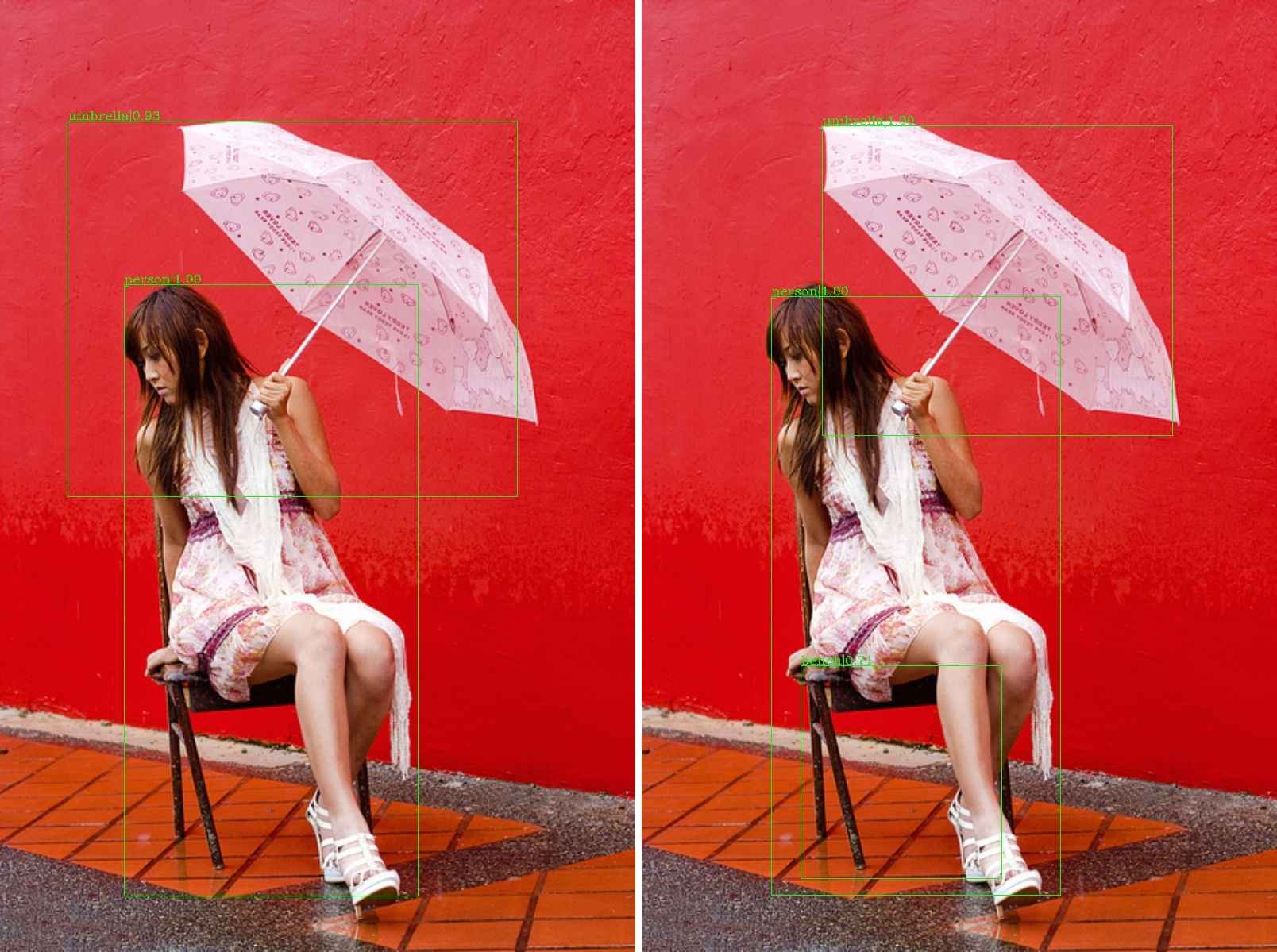}
		\includegraphics[width=.68\linewidth]{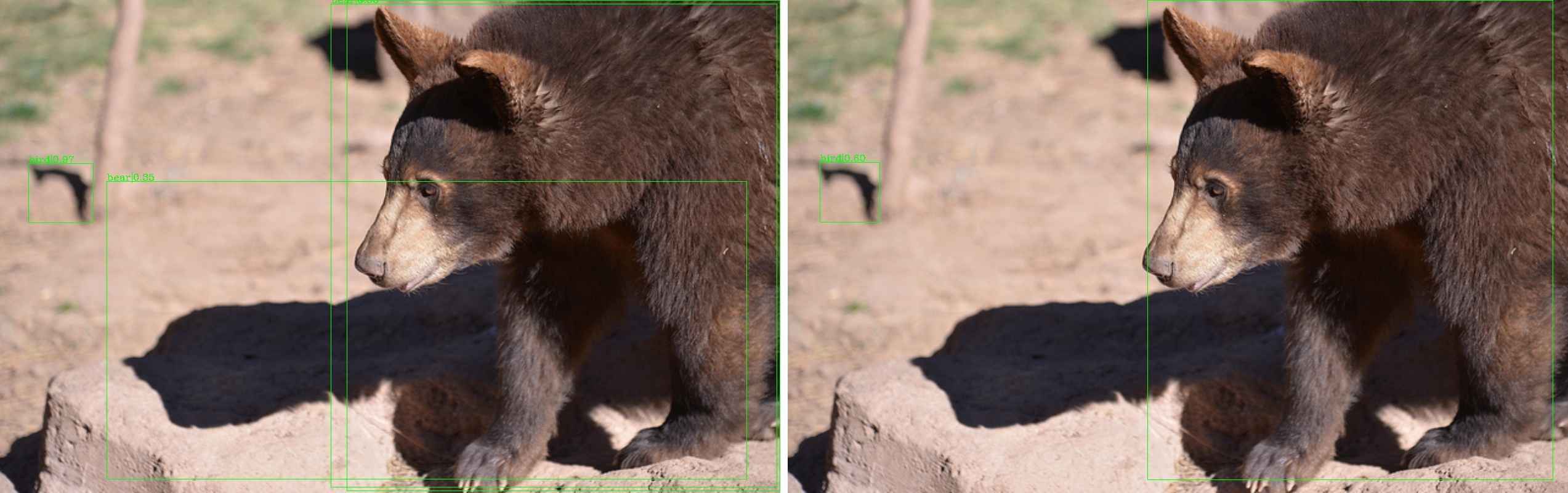}
	\end{center}
	\caption{Some bounding-box detection results of Faster R-CNN with ResNet-101 on the COCO 2017 validation set.}
	\label{fig:frcnn-2}
\end{figure}

\begin{figure}[htbp]
	\begin{center}
		\includegraphics[width=.8\linewidth]{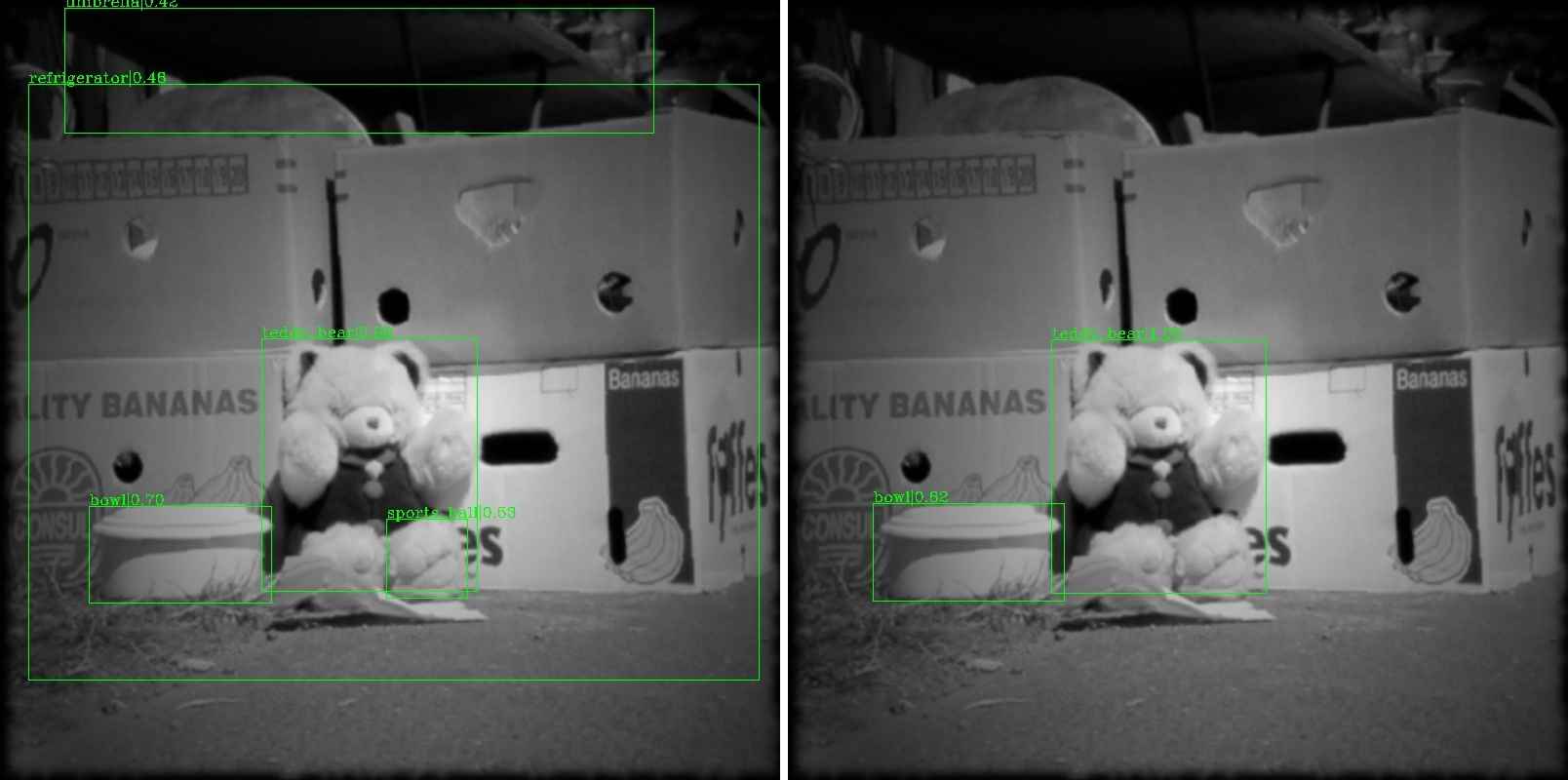}
		\includegraphics[width=.8\linewidth]{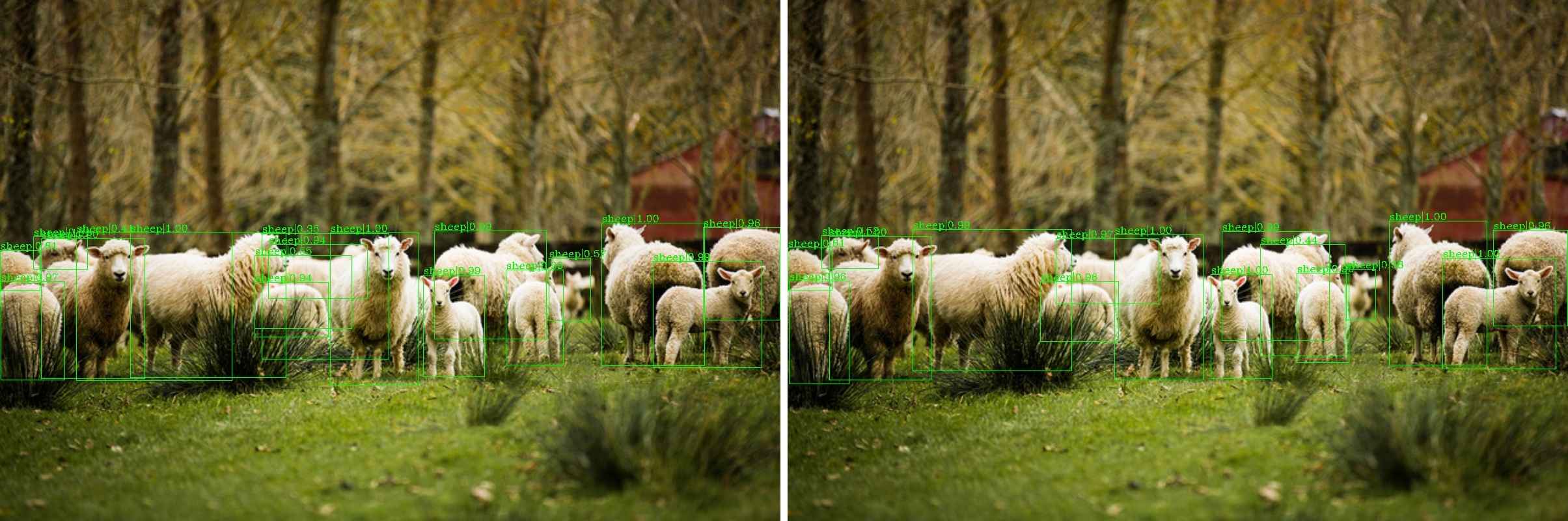}
		\includegraphics[width=.8\linewidth]{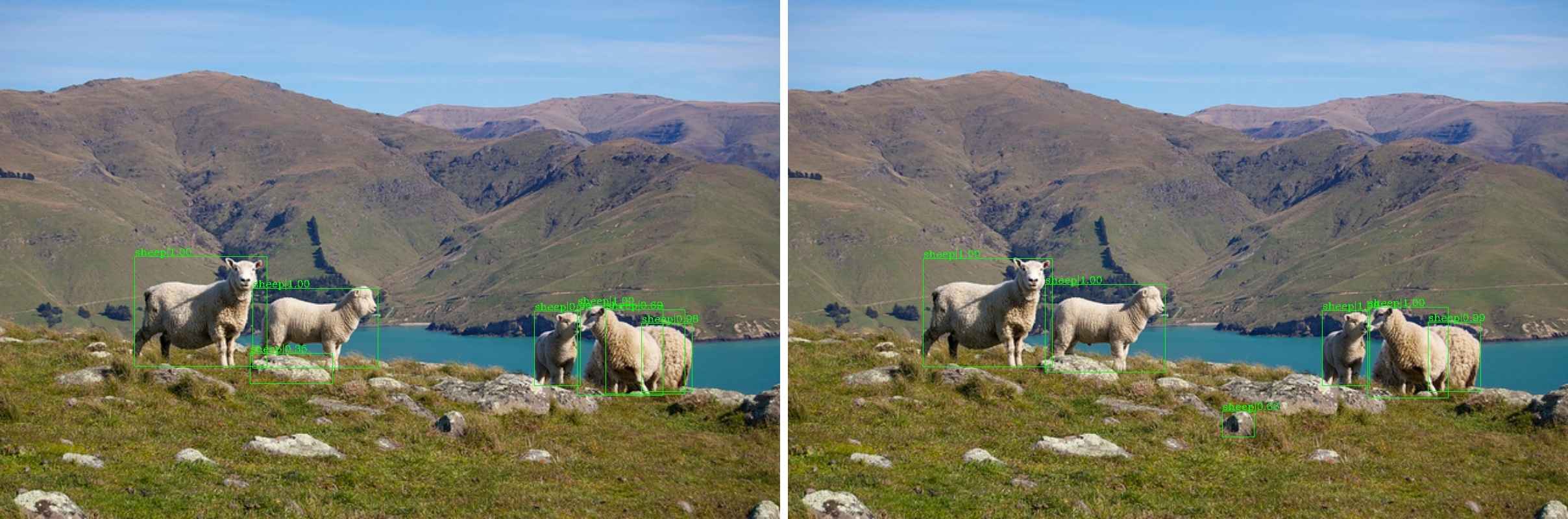}
		\includegraphics[width=.8\linewidth]{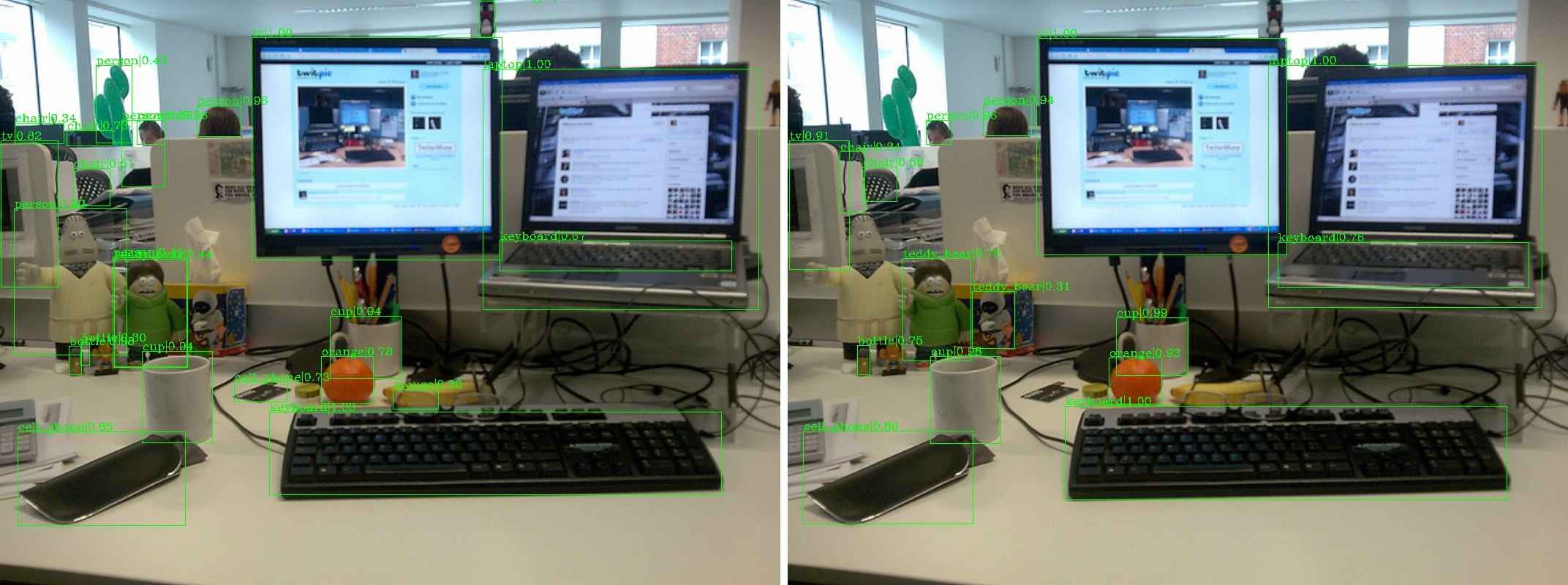}
	\end{center}
	\caption{Some bounding-box detection results of Faster R-CNN with ResNet-101 on the COCO 2017 validation set.}
	\label{fig:frcnn-3}
\end{figure}

The result comparisons of Faster R-CNN are presented in Fig.~\ref{fig:frcnn-1}, \ref{fig:frcnn-2} and \ref{fig:frcnn-3}. Regarding the bounding-box results, detectors based on PSConv could reduce false alarms of large-sized objects and precisely perceive small-sized instances. For example, the potted plant in the second row of Fig.~\ref{fig:frcnn-1}, the bear in the last row of Fig.~\ref{fig:frcnn-2}, the refrigerator in the first row of Fig.~\ref{fig:frcnn-3} are obvious false alarms that are rejected in the predictions of our model. Furthermore, referring to the middle row in Fig.~\ref{fig:frcnn-2}, the bounding box of the umbrella is more compact and the bench below the person is detected with confidence. It validates the superiority of our PSConv-based detector to capture objects with diverse shapes and sizes.

\begin{figure}[htbp]
	\begin{center}
		\includegraphics[width=.85\linewidth]{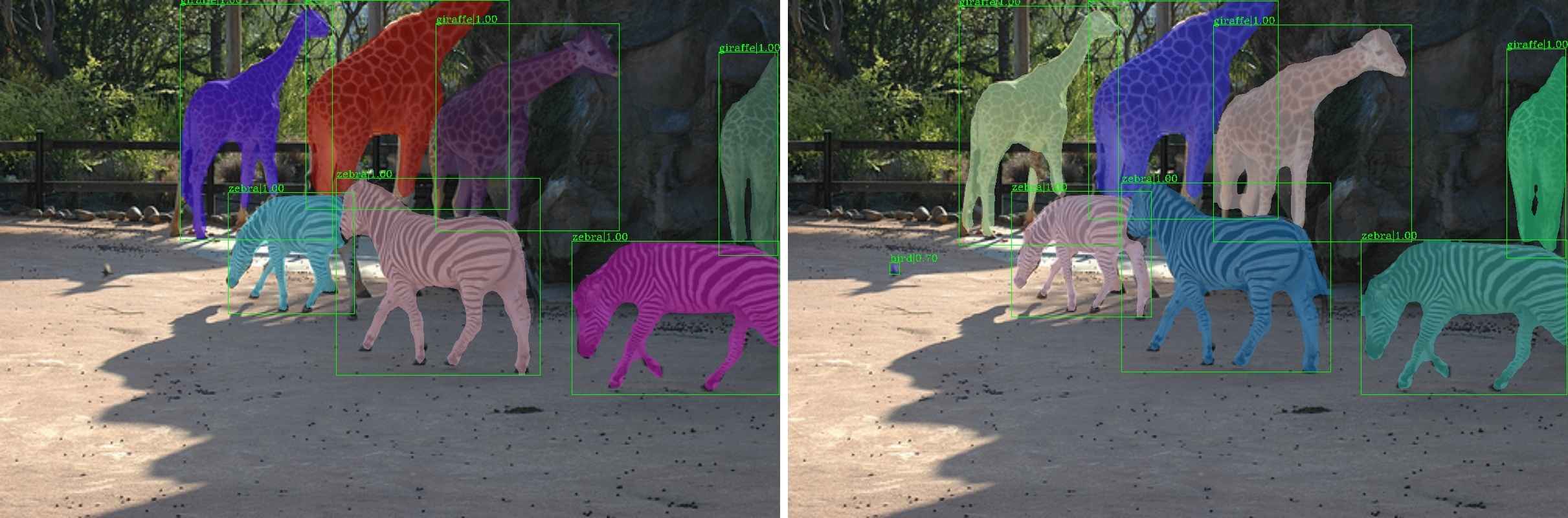}
		\includegraphics[width=.85\linewidth]{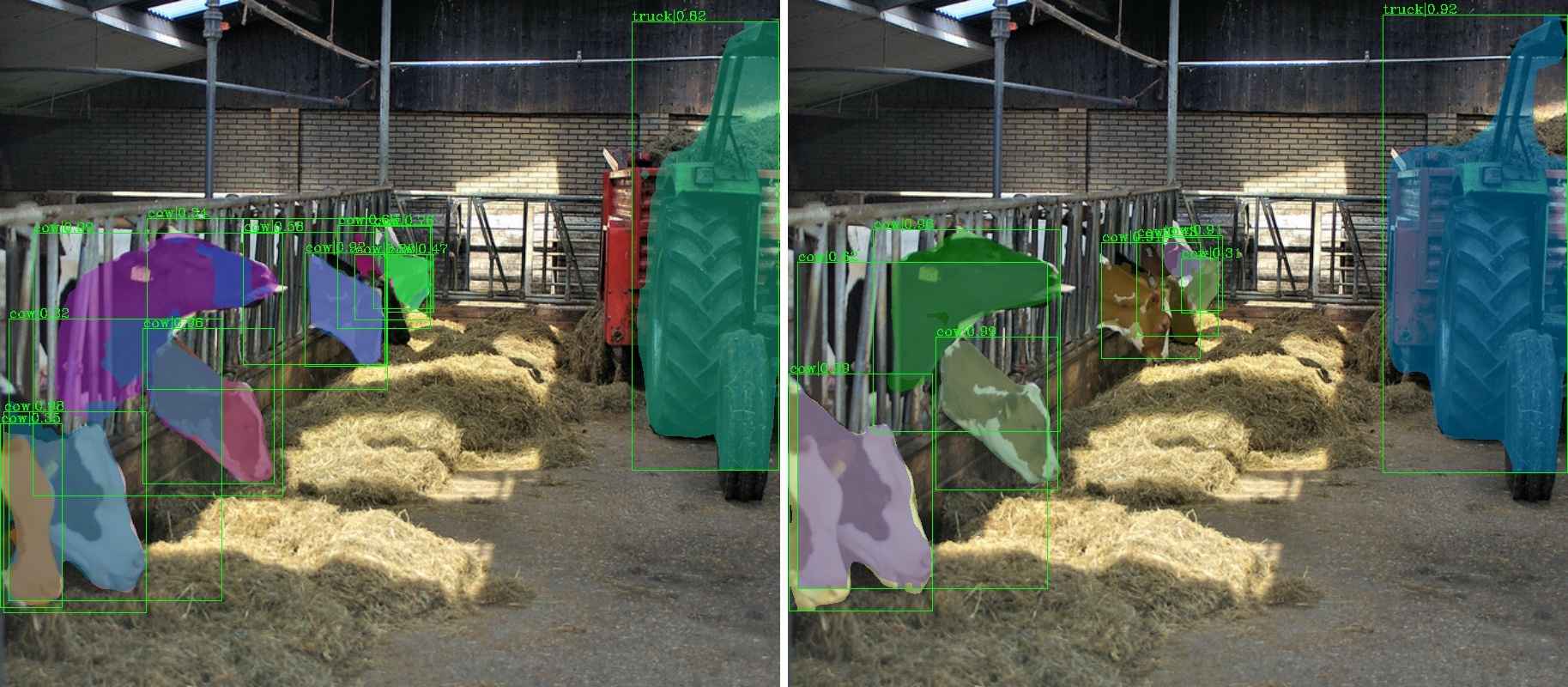}
		\includegraphics[width=.85\linewidth]{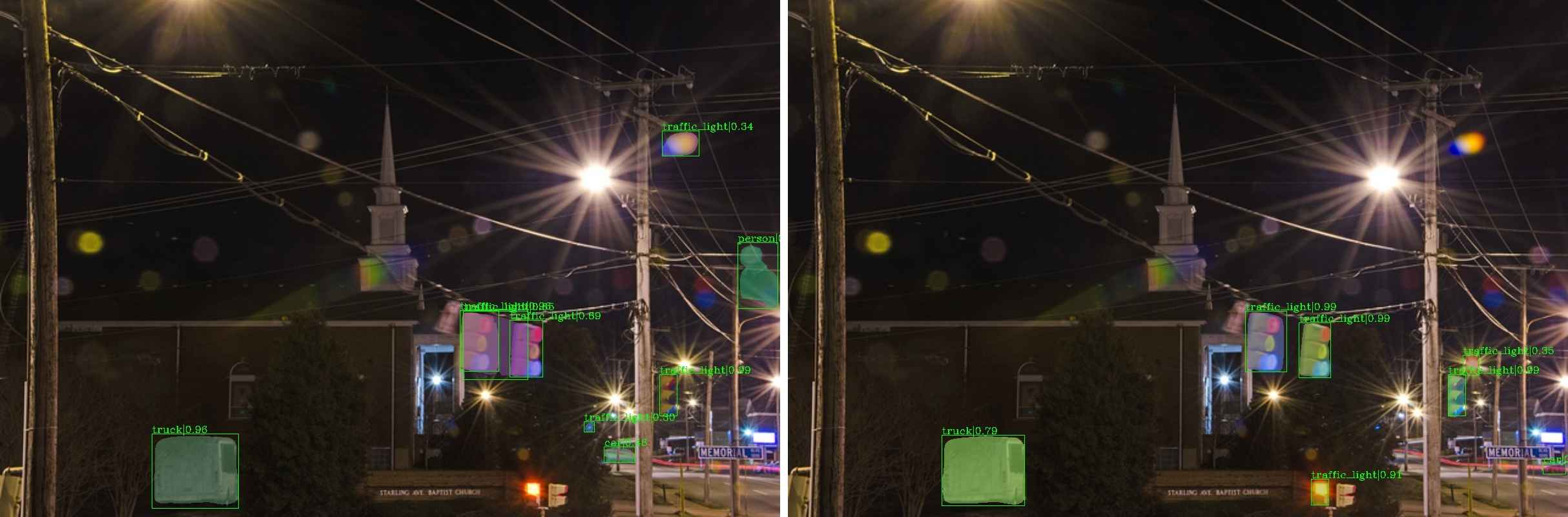}
		\includegraphics[width=.85\linewidth]{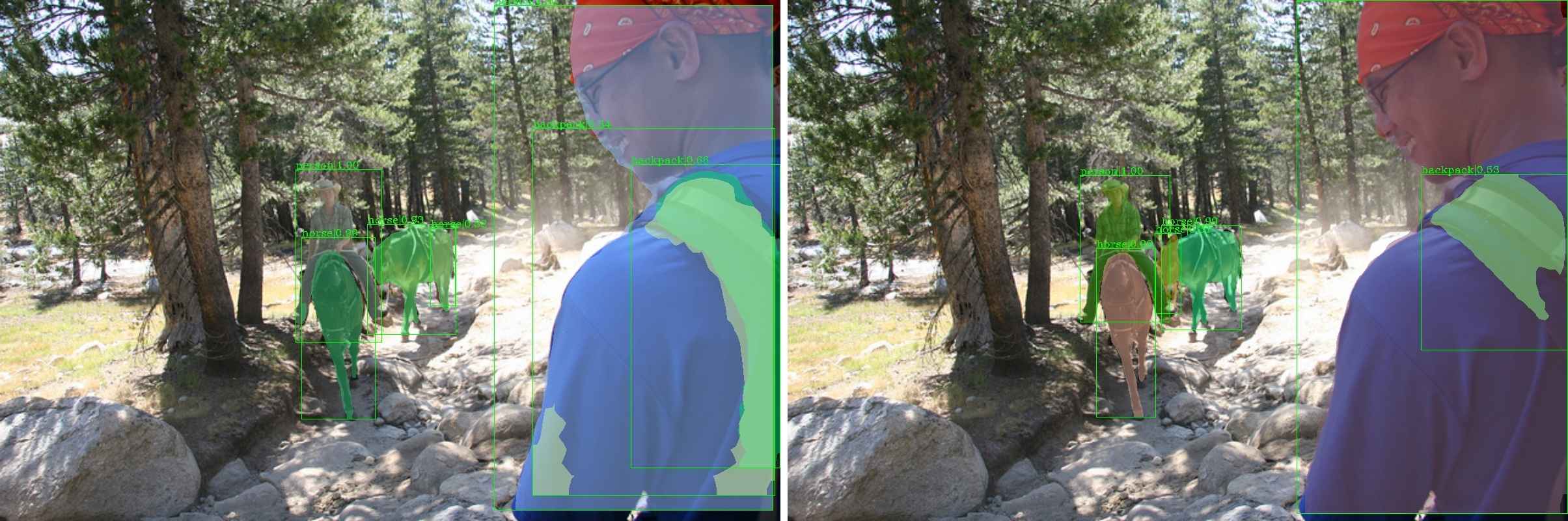}
	\end{center}
	\caption{Some bounding-box detection and instance segmentation results of Mask R-CNN with ResNet-101 on the COCO 2017 validation set.}
	\label{fig:mrcnn-1}
\end{figure}

\begin{figure}[htbp]
	\begin{center}
		\includegraphics[width=.85\linewidth]{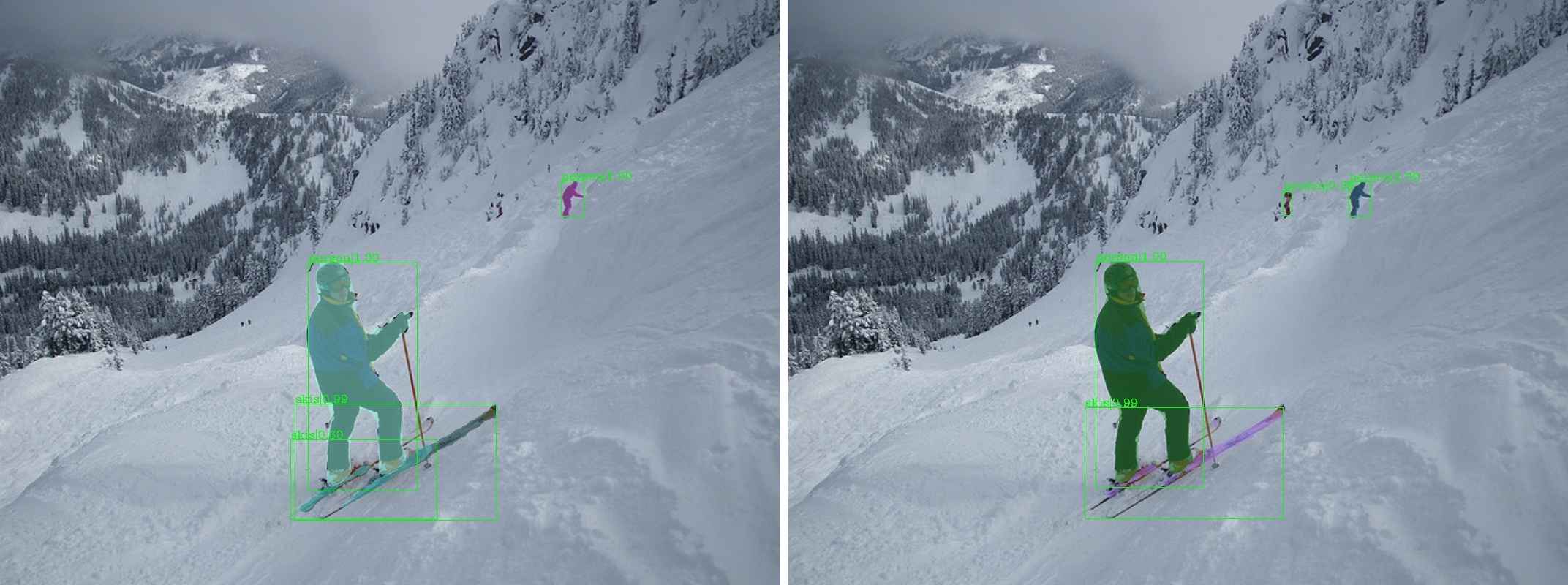}
		\includegraphics[width=.85\linewidth]{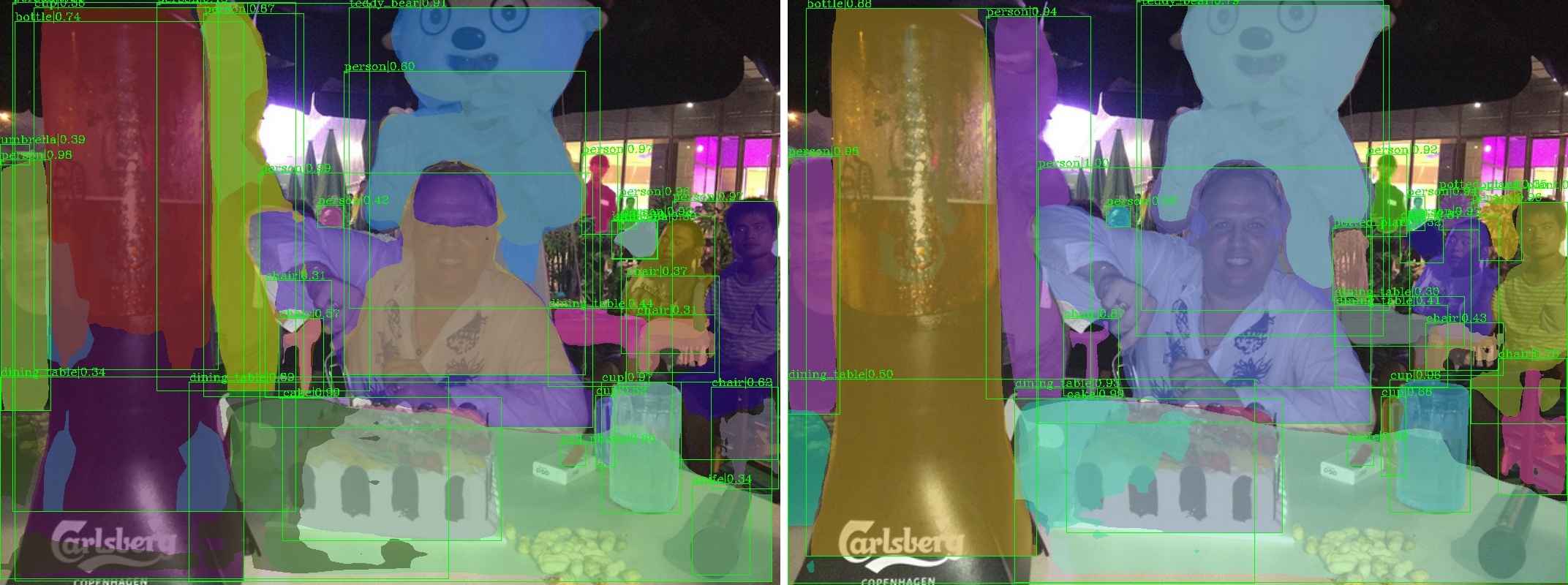}
		\includegraphics[width=.85\linewidth]{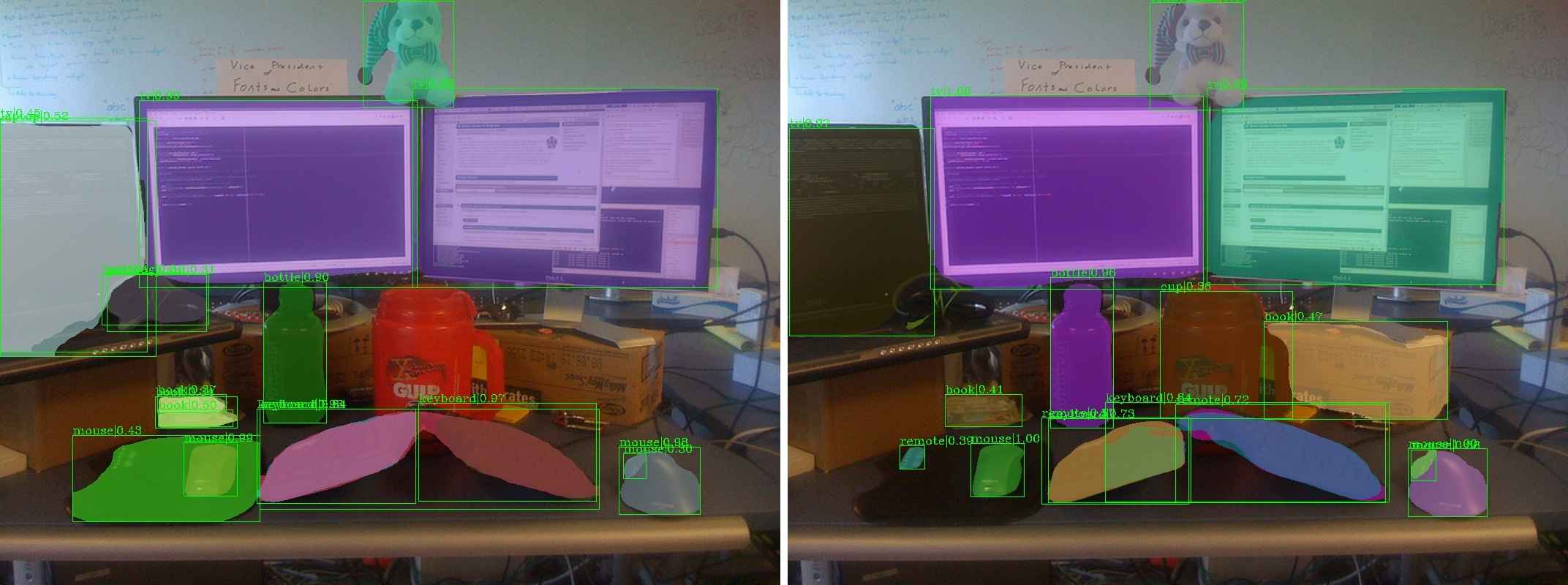}
		\includegraphics[width=.85\linewidth]{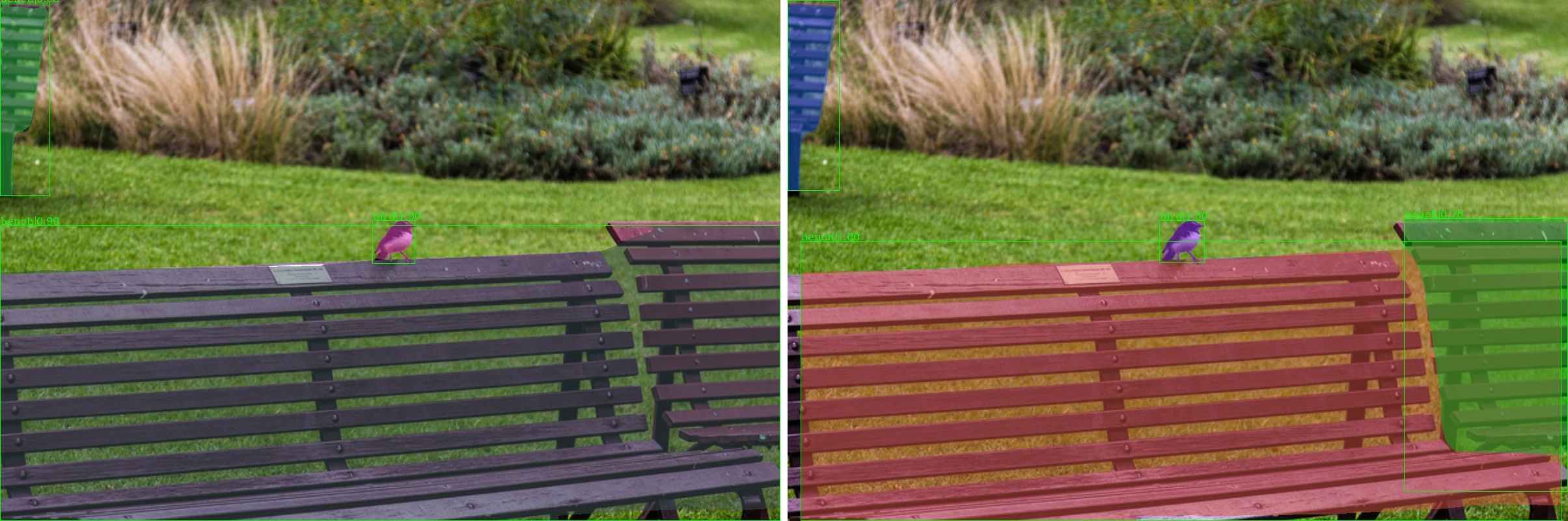}
	\end{center}
	\caption{Some bounding-box detection and instance segmentation results of Mask R-CNN with ResNet-101 on the COCO 2017 validation set.}
	\label{fig:mrcnn-2}
\end{figure}

\begin{figure}[htbp]
	\begin{center}
		\includegraphics[width=.93\linewidth]{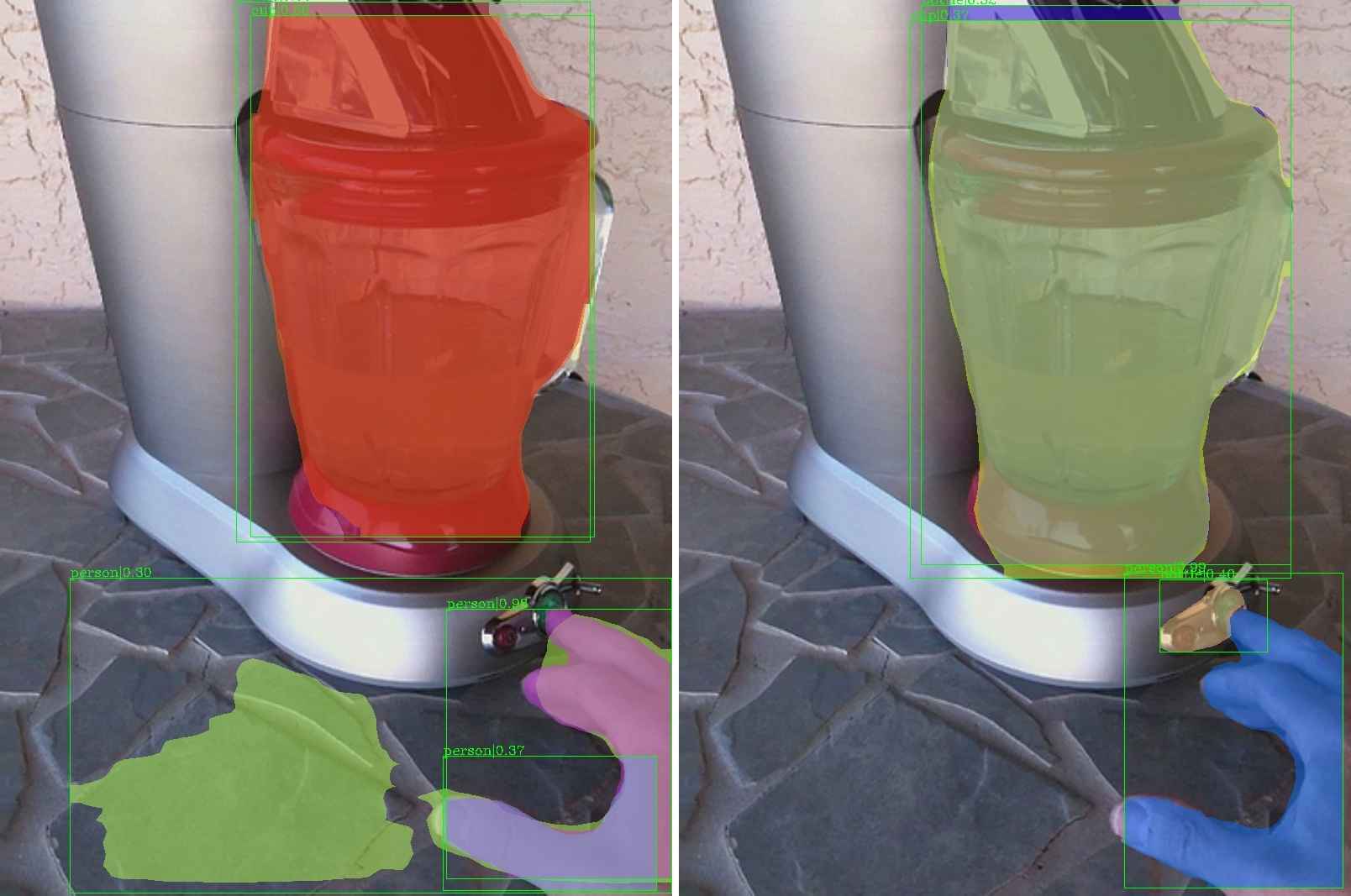}
		\includegraphics[width=.93\linewidth]{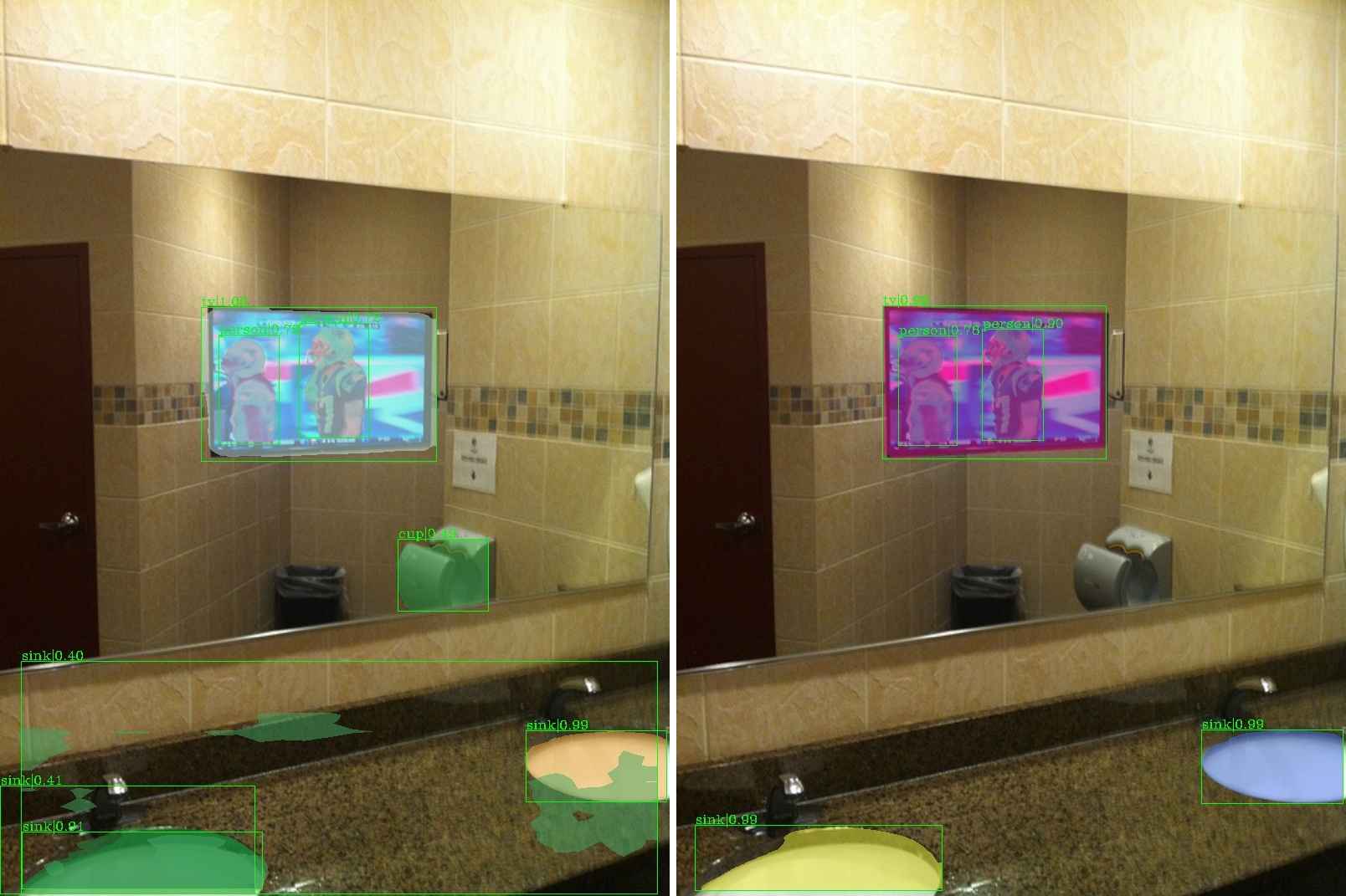}
	\end{center}
	\caption{Some bounding-box detection and instance segmentation results of Mask R-CNN with ResNet-101 on the COCO 2017 validation set.}
	\label{fig:mrcnn-3}
\end{figure}

The result comparisons of Mask R-CNN are presented in Fig.~\ref{fig:mrcnn-1}, \ref{fig:mrcnn-2} and \ref{fig:mrcnn-3}. For example, the traffic light in the third row of Fig.~\ref{fig:mrcnn-1}, the person in the first row and the sink in the last row of Fig.~\ref{fig:mrcnn-3} are false alarms in the standard detector but omitted in our PSConv-based detector. A skiing person on the snow mountain is missed by the standard detector possibly due to its tiny size, but successfully detected by the PSConv-based model, as shown in the first row of Fig.~\ref{fig:mrcnn-2}. As demonstrated in the last row of Fig.~\ref{fig:mrcnn-2}, distinct instances of the bench are distinguished together with the detected small bird, thanks to the robustness of our PSConv to scale variation.

\section{Speed Optimization}
Based on our preliminary GPU speed benchmark in the main paper, the speed gap is primarily due to dilated convolution inside our PSConv. However, such a gap can be largely bridged using a specialized implementation of Dilated-Winograd Convolution (DWC). Compared to GEMM implementation in cuDNN, the average speedup by DWC for dilated convolutional layers with a dilation rate of 2/4 is $2.14\times$/$1.53\times$, on a single TITAN X GPU (similar results are also reported in~\cite{8803277}). By adopting the TVM compiler~\cite{222575}, the speedup can be further increased to $2.86\times$/$2.01\times$. After combining the latest version of Intel OneDNN tool (achieving an additional speedup of +0.2), the inference time of a PSConv layer would be roughly $1.42\times$ of the standard convolution. Furthermore, the above optimization procedure could yield a better speedup ratio on CPU inference, tested on Dual Intel Xeon Platinum 8280 @ 2.70GHz. Since we only apply PSConv to the $3 \times 3$ convolutional layers of a residual network, the slow-down effect will be diluted on a whole network compared to a single convolution layer. Specifically, the inference time of a PSConv-based ResNet-50/101 becomes very similar to the standard ResNet-50/101 ($1.066\times$ on GPU and $1.051\times$ on CPU). As a consequence, our PSConv can be comfortably put into practical usage.

%
%
\bibliographystyle{splncs04}
\bibliography{egbib}
\end{document}